\pdfoutput=1

\documentclass[11pt]{article}

\usepackage[]{ACL2023}

\usepackage{times}
\usepackage{latexsym}

\usepackage[T1]{fontenc}

\usepackage[utf8]{inputenc}

\usepackage{microtype}

\usepackage{inconsolata}

\usepackage{multirow}
\usepackage{amsmath}
\usepackage{cleveref}
\usepackage{tcolorbox}
\usepackage{xcolor,colortbl}
\usepackage{caption}
\usepackage{booktabs,subcaption,amsfonts,dcolumn}
\usepackage{pifont}
\usepackage{ulem} 
\usepackage{tikz}
\usepackage{bbding}
\usepackage{cancel}
\usepackage{graphicx}
\usepackage{hyperref}  
\usepackage{enumitem}
\usepackage{algorithm}
\usepackage{algpseudocode}

\newtcolorbox{mybox}[1]{colback=gray!10!white,colframe=gray!75!black, boxrule=0.1mm,fonttitle=\bfseries,title=#1}
\newcommand{\grayc}[0]{\cellcolor[rgb]{0.957,0.957,0.957}}
\newcommand{\g}[0]{\color{gray!50}}

\definecolor{GREEN}{HTML}{BEE7A5} 
\definecolor{PINK}{HTML}{FFC0CB} 

\newcommand{\ck}[1]{%
  \ifnum#1=6 \cellcolor{GREEN!100!white}#1%
  \else\ifnum#1=5 \cellcolor{GREEN!66!white}#1%
  \else\ifnum#1=4 \cellcolor{GREEN!33!white}#1%
  \else\ifnum#1=3 \cellcolor{PINK!25!white}#1%
  \else\ifnum#1=2 \cellcolor{PINK!50!white}#1%
  \else\ifnum#1=1 \cellcolor{PINK!75!white}#1%
  \else \cellcolor{PINK!100!white}#1%
  \fi\fi\fi\fi\fi\fi%
}

\newcommand{\cd}[1]{%
 \ifnum#1=4 \cellcolor{GREEN!100!white}#1%
  \else\ifnum#1=3 \cellcolor{GREEN!50!white}#1%
  \else\ifnum#1=2 \cellcolor{PINK!33!white}#1%
  \else\ifnum#1=1 \cellcolor{PINK!66!white}#1%
  \else \cellcolor{PINK!100!white}#1%
  \fi\fi\fi\fi
}

\algrenewcommand\alglinenumber[1]{}

\title{Are Human \underline{\textit{Interactions}} Replicable by Generative Agents?
\\A Case Study on Pronoun Usage in Hierarchical Interactions}

\author{Naihao Deng \and Rada Mihalcea\\
    University of Michigan\\
    $^{2}$\texttt{\{dnaihao, mihalcea\}@umich.edu}}

\begin{document}
\maketitle
\begin{abstract}
Recently, researchers have increasingly employed Large Language Models (LLMs) for social simulations.
In this paper, we explore whether \textit{interactions} among LLM agents exhibit human-like behaviors, with a specific focus on pronoun usage differences between leaders and non-leaders. 
Our evaluation uncovers significant discrepancies that neither prompt-based configurations nor specialized agent setups successfully replicate human-like pronoun usage patterns. 
Furthermore, we demonstrate that even when LLMs possess an understanding of these patterns, they fail to exhibit them in their interactions.
Our study highlights the limitations of LLM-based social simulations and underscores the need for caution when relying on such simulations for understanding or modeling human social behaviors.
\end{abstract}

\section{Introduction}

As Large Language Models (LLMs) become increasingly powerful \citep{radford2019language, touvron2023llama, dubey2024llama}, researchers have used them to simulate human behavior and social interactions \citep{park2022social}, and demonstrate that these LLMs can produce simulacra of both individual and emergent group behavior \citep{park2023generative}.
This line of research has inspired numerous follow-up studies exploring the use of LLMs for role-playing \citep{wang-etal-2024-rolellm}, and social prototyping in virtual worlds and games \citep{treanor2024prototyping}.
Moreover, researchers have leveraged such simulations to address social issues \citep{piatti2024cooperate} and simulate human interactions \citep{gao2023s, hua2023war, li-etal-2024-econagent}, hoping to facilitate decision-makers to make informed decisions, prevent harms, etc.

Despite this promise, questions remain about the extent to which LLMs authentically resemble human behaviors \citep{jiang2023personallm}.
While there is an ongoing debate, most studies are conducted by researchers prompting LLMs and examining their responses.
In contrast, there is a paucity of research examining how well LLMs simulate human-to-human interactions in multi-agent settings.
Evaluating these group-level interactions is critical, as societal decision-making relies not only on the outcomes of the interactions but also on the processes that lead to those decisions.
Prior studies have highlighted the emergence of behaviors in LLMs that mimic human traits \citep{park2023generative}, but the fidelity of these behaviors in capturing the nuances of human interaction remains underexplored.
We argue that understanding these nuanced interaction patterns is crucial to ensure that LLM-driven simulations accurately reflect the human interaction process.
Otherwise, the conclusions drawn from such simulations may be misleading and fail to accurately reflect real-world dynamics.

\begin{figure*}[t]
    \centering
    \includegraphics[width=0.9\linewidth]{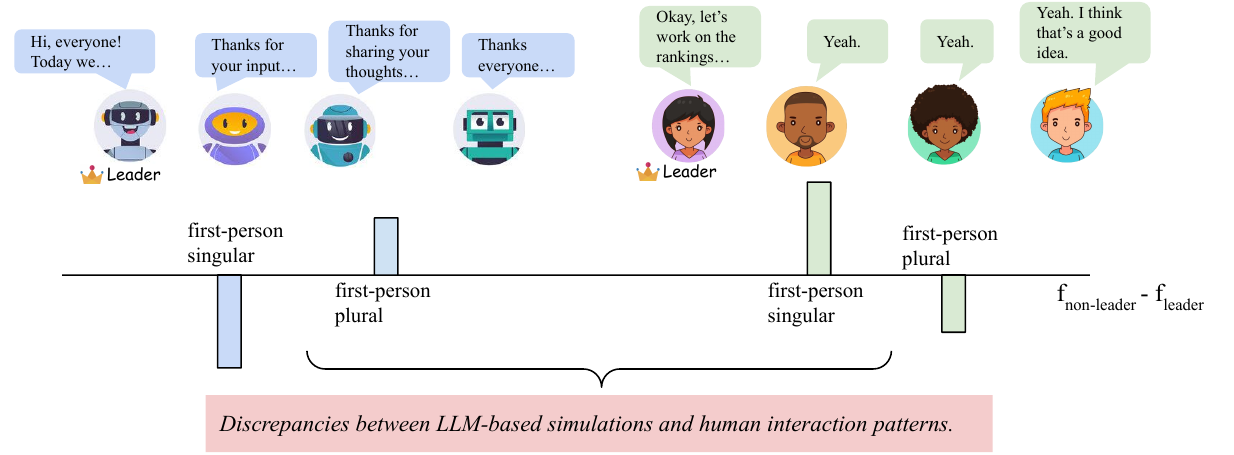}
    \caption{
    We follow \citet{kacewicz2014pronoun}'s setup but replace human subjects with LLM agents.
    ``f'' represents the frequency corresponding to each pronoun type.
    We reveal that contrary to human results by \citet{pennycook1994politics, kacewicz2014pronoun}, most ``non-leader'' LLMs do not use first-person singular pronouns more often (f\textsubscript{non-leader}-f\textsubscript{leader} < 0), and ``leader'' LLMs do not use first-person plural pronouns more often in their interactions (f\textsubscript{non-leader}-f\textsubscript{leader} > 0).
    We present additional transcripts in \Cref{app-sec:dialogue-example}.
    }
    \label{fig:concept-diagram}
\end{figure*}


In this paper, we explore how closely interactions between LLM agents mirror those of human beings.
Specifically, we study the difference in pronoun usage between leaders and non-leaders during interactions (shown in \Cref{fig:concept-diagram}) as pronoun usage is embedded in human communication, reflecting hierarchical roles, self-perception, and relational attitudes \citep{brewer1996we, stets2003bringing, campbell2003secret, pennebaker2011secret}.
Pronoun usage patterns during interactions cannot be consciously controlled over extended periods, making them a natural indicator of underlying behavioral tendencies \citep{mehl2006personality, ireland2014natural}.
Such unconscious human patterns can pose significant challenges for LLMs to replicate.
In addition, there are extensive efforts from psychology revealing the consistent effect on the use of pronouns \citep{kacewicz2014pronoun} and the reason behind \citep{duval1972theory, davis1975use} among human beings.


We adopt \citet{kacewicz2014pronoun}'s setup while replacing the human subjects with LLM agents.  
We evaluate a diverse set of LLMs, including those from GPT families \citep{achiam2023gpt}, Llama-3.1 families \citep{dubey2024llama}, Mistral families \citep{jiang2023mistral}, and QWen families \citep{bai2023qwen}.
Our simulation involves various persona prompts employed in the existing literature.
In addition, we evaluate a specialized agent that incorporates reflection and planning proposed by \citet{park2023generative}.

Our results reveal the significant discrepancies in pronoun usage between LLM-based simulations and human interactions.
Contrary to human results by \citet{pennycook1994politics, kacewicz2014pronoun}, most ``non-leader'' LLMs do not use first-person singular pronouns more often, and ``leader'' LLMs do not use first-person plural pronouns more often.
We argue that if LLMs fail to replicate these subtle yet critical markers, it raises concerns about their ability to authentically model complex social dynamics.
Despite the excitement brought by the LLM agents, we advocate practitioners to understand the limitations of using LLMs for social simulation.

Our paper makes three main contributions.
First, to the best of our knowledge, we are the first to study whether LLM agents demonstrate human-like leader and non-leader pronoun usage patterns in their interactions.
Second, we evaluate LLM agents from different model families, various persona prompts from existing literature, and agent designs. 
Finally, we highlight that in our simulation, LLM agents barely resemble human-like pronoun usage in their interaction process, even if the LLM agent shows some level of understanding of the relevant knowledge. 
Therefore, we strongly suggest practitioners realize the limitations of the social simulation based on LLM agents.

\section{Related Work}

\paragraph{Persona LLMs.}
Prior research suggests that when LLMs receive different inputs, varying parts of their parameters activate, leading to shifts in behavior \citep{zhang-etal-2022-moefication, wang2024exploring}. 
Studies have shown that LLM behaviors are highly influenced by prompts across diverse NLP tasks, highlighting the potential of prompt engineering to shape LLM behavior \citep{white2023prompt, chang2023prompt, deng-etal-2024-tables}.
Existing work has evaluated how well persona-based LLMs mimic human behaviors. 
For instance, \citet{aher2023using} examine the limitations of persona LLMs but focus primarily on classification tasks, overlooking multi-agent interaction. 
Similarly, \citet{li2024measuring} identify persona decay over extended dialogues, revealing challenges in maintaining persona consistency.
From an application perspective, \citet{xu2023expertprompting, wang2023can} demonstrate that assigning expert personas improves LLM performance on downstream tasks. 
\citet{salewski2024context} explore LLMs' strengths and underlying biases by leveraging their role-playing ability.

\paragraph{Social Simulation.}
Social simulation is a natural extension of persona-based LLMs. 
By placing various persona LLMs in simulated environments, researchers can study interaction patterns within these frameworks \citep{kaiya2023lyfe}. 
Such multi-agent systems have been employed to explore topics like social networks \citep{gao2023s}, macroeconomic activities \citep{li-etal-2024-econagent}, job-seeking behaviors \citep{li2023metaagents}, decisions and consequences in wars \citep{hua2023war}, and even interactions between humans and extraterrestrial civilizations \citep{jin2024if}.
While these studies demonstrate the potential of multi-agent systems to yield valuable insights, they often fail to address the reliability of these simulations, particularly when modeling human interactions. 
In this paper, we aim to uncover the limitations of such simulations from the aspect of pronoun usage patterns.

\section{Simulation Setups}

We introduce the persona prompts, how we construct the specialized agent, the simulated scenarios following \cite{kacewicz2014pronoun}, and other considerations in this section.

\begin{table}[t]
    \centering
    \small
    \begin{tabular}{p{2.3cm}p{4.5cm}}
    \toprule
       Source  &  Persona Prompts\\
       \midrule
       \multirow{2}{2.3cm}{\citet{park2023generative}} & \grayc Name: \{\textsc{Name}\} (age: \{\textsc{Age}\}) Innate traits: \{\textsc{Persona}\}\\
       \midrule
       \multirow{4}{2.3cm}{\citet{deshpande-etal-2023-toxicity}}  & \grayc Speak exactly
like \{\textsc{Persona}\}. 
        Your response should copy the style of \{\textsc{Persona}\},
both the writing style and words you use. \\
    \midrule
   \multirow{4}{2.3cm}{\citet{gupta2023bias}}  & \grayc Adopt the identity of \{\textsc{Persona}\}. Respond while staying in strict accordance with the nature of this identity. \\
   \midrule
    \multirow{4}{2.3cm}{\citet{borah2024implicitbiasdetectionmitigation}} & \grayc Assume you are \{\textsc{Persona}\}. Do not explicitly say: `as an AI model'. Stick to your roles and do not share you are an AI or an AI agent. \\
    \bottomrule
    \end{tabular}
    \caption{Prompts used in existing literature to elicit persona in LLMs.
    We evaluate LLMs with each of these prompts in \Cref{sec: results}, denoted as P1 to P4 accordingly.
    }
    \label{tab:persona-prompts}
\end{table}

\subsection{Persona Prompts}
\label{subsec: persona-prompts}

\Cref{tab:persona-prompts} provides the persona templates used in the existing literature.
In \Cref{sec: results}, we test each of them denoted as P1 to P4 on a variety of LLMs.

\subsection{LLM Agent}
\label{subsec: llm-agent}

\paragraph{Simple Agent.}
The agent incorporates every sentence in the history and these sentences are passed in as the context.

\[
    R = LLM(P \oplus H),
\]

\noindent where $H = \{u_1: s_1, u_2: s_2, \cdots, u_i: s_i \cdots \}$ represents the history where the agent $u_i$ utters the sentence $s_i$.
$P$ is the persona prompts (e.g. P1 to P4 from \Cref{subsec: persona-prompts}).
$\oplus$ indicates concatenation.
$R$ is the LLM's response.
We truncate the history when it is longer than the LLM's context window.

\paragraph{Specialized Agent.}
\citet{park2023generative} propose the generative agent architecture which involves memory stream, reflection, planning, memory retrieval, etc.
Similarly, we adapt the architecture here to our scenario.
Our framework encompasses several key components:

\noindent1. \textit{Memory.} Our memory stream maintains a record of all the discussion history, $H = \{u_1: s_1, u_2: s_2, \cdots, u_i: s_i \cdots \}$.
Since the study we simulate lasts 30 minutes per group, and all LLMs' responses are based on the interactions among these LLMs, we do not incorporate a retrieval component in our framework.
We truncate the history when the history is longer than the LLM's context window.

\noindent2. \textit{Reflection.} Reflections are abstract thoughts generated by the agent.
We incorporate such a reflection process when one round of a conversation finishes.
We adapt the original prompt from \citet{park2023generative} as follows:

\begin{mybox}{Reflection Prompt}
    Given the conversation history, reflect at high-level about the current situation in three sentences.
\end{mybox}

\noindent3. \textit{Planning.}
We incorporate a \textit{planning} session before each agent starts to speak as follows: 
\begin{mybox}{Planning Prompt}
    Given the conversation history, plan your speech at high-level in three sentences.
\end{mybox}

\subsection{Simulated Scenario: Task-Oriented Conversation}
\label{subsec: scenario-setup}
We mimic the setup from the real-world human experiments conducted by \citet{burris2009playing, kacewicz2014pronoun}.
We randomly assign an LLM agent as the leader, therefore a four-agent group comprises exactly one leader and three subordinates consistent with \citep{burris2009playing, kacewicz2014pronoun}'s group setup.
The agents first work on their own and then discuss the rankings as a team.
We repeat the group experiments 41 times as there are 41 four-person groups in the studies by \citet{burris2009playing, kacewicz2014pronoun}.
In addition, we follow the original demographic statistics to assign the demographic background to each LLM agent, such as gender.
\Cref{alg: simulation} presents an overview of the simulation process.
\Cref{app-sec: additional-experimental-setups} provides additional details of our experimental setups.

\begin{algorithm}[t]
\caption{Simulation: Task-Oriented Conversation}
\label{alg: simulation}
\begin{algorithmic}[1]
  \State \texttt{num\_groups} $\leftarrow 41$, \texttt{turns} $\leftarrow 3$
\For{j from 1 to \texttt{num\_groups}}
\For{i from 1 to 4}
    \State Agent\textsubscript{i} = Assign(role, demographics)
    \State Agent\textsubscript{i}.individual\_ranking()
\EndFor        

\For{j from 1 to \texttt{turns}}
\For{i from 1 to 4}
    \State Agent\textsubscript{i}.speak()
    \State Agent\textsubscript{$k \in \{1, 2, 3, 4\}\setminus i$}.listen()
\EndFor  
\EndFor  
\EndFor
\end{algorithmic}
\end{algorithm}


\subsection{Other Considerations}

\paragraph{Orders of Speaking.}
In real-life human interactions, there are natural interruptions, leading to variants in speaking orders.
Some prior works adopt a controller to determine the speaking order to mitigate such issues \citep{zhang2024simulating}.
However, as we study the frequency of pronoun usages in \Cref{sec: results}, we simplify the setup and use a round-robin fashion for each agent to speak up.

\paragraph{Rounds of Interactions.}
The four-person group in the original study interacts for 30 minutes \citet{kacewicz2014pronoun}. 
As the average speaking speed of a person is around 130 words per minute\footnote{\url{https://www.typingmaster.com/speech-speed-test/\#:~:text=In\%20general\%2C\%20it\%20is\%20considered,or\%20reading\%20aloud\%20a\%20passage.}}, and three rounds of interactions lead to around 4,000 words through preliminary experiments, we make LLMs interact for three rounds in our experiments.
In addition, as our experiments focus on comparing whether the leader or the non-leader uses certain pronouns at a higher frequency, the frequency does not rely on the number of words generated from each agent.
Therefore, the rounds of interactions would not influence our findings.

\section{Experimental Setup}

\subsection{Models}

We test a series of closed-source GPT models from OpenAI, including \textbf{GPT-3.5-Turbo} at its June 13th version, \textbf{GPT-4-Turbo} at its June 13th version, and \textbf{GPT-4o} at its May 13th 2025 version \citep{achiam2023gpt}.
We test a series of open-source LLMs from the Llama 3.1 family, including \textbf{Llama-3.1-8B-Instruct}, \textbf{Llama-3.1-70B-Instruct}, and \textbf{Llama-3.1-405B-Instruct} \citep{dubey2024llama}.
For the \textbf{Llama-3.1-405B-Instruct} model, we deploy the GPTQ quantized \citep{frantar2022gptq} version of the model at INT4.
We test the largest Mistral model, \textbf{Mistral-Large-Instruct-2407} \citep{jiang2023mistral} published in July 2024.
For consistency, we denote this model as Mistral 123B Instruct in the paper.
We test the QWen 2.5 models from \textbf{QWen-2.5-0.5B-Instruct} to \textbf{QWen-2.5-72B-Instruct} model \citep{bai2023qwen} published in September 2024.  

For the open-source LLMs, we deploy them locally on a single server with eight A100 GPUs.
We accelerate the model inference by using the VLLM \citep{kwon2023efficient} library.

\subsection{Evaluation}
We examine the subtle yet critical marker in human {\it interaction} -- the pronoun usage.
Specifically, in our simulation for task-oriented conversations, we aim to determine if the pronoun usage patterns of LLMs align with those observed in well-studied human interactions. 
Prior research in psychology has shown that non-leaders tend to use first-person singular pronouns (e.g., "I," "me") more frequently than leaders, while leaders tend to use first-person plural pronouns (e.g., "we," "us") more frequently \citep{kacewicz2014pronoun}. 
Therefore, we analyze whether there is a statistically significant difference in pronoun usage between non-leader and leader agents that aligns with human patterns. 
For example, in \Cref{fig: simple-agent-advanced-models}, the bars correspond to the difference, $\Delta$, between the average first-person singular pronoun use frequency of non-leaders and leaders, concretely,
\begin{equation}\label{eq: eval-calc}
    \Delta = f_{\text{non-leaders}, \text{avg}} - f_{\text{leaders}, \text{avg}},
\end{equation}
where $f_{\text{avg}}$ represents the average frequency of pronoun usage expressed as a percentage value, $f\in[0, 100]$.
In \Cref{fig:first-person-singular}, the bars representing GPT-4o with Prompt 2, 3, and 4 exhibit a similar pattern to that of humans, all showing statistically significant positive results. 
\Cref{app-sec: additional-experimental-setups} provides additional details on how we conduct the statistical significance test.


\section{Findings}
\label{sec: results}

\begin{figure}[t]
\centering
    \begin{subfigure}{\linewidth}
        \includegraphics[width=0.95\linewidth]{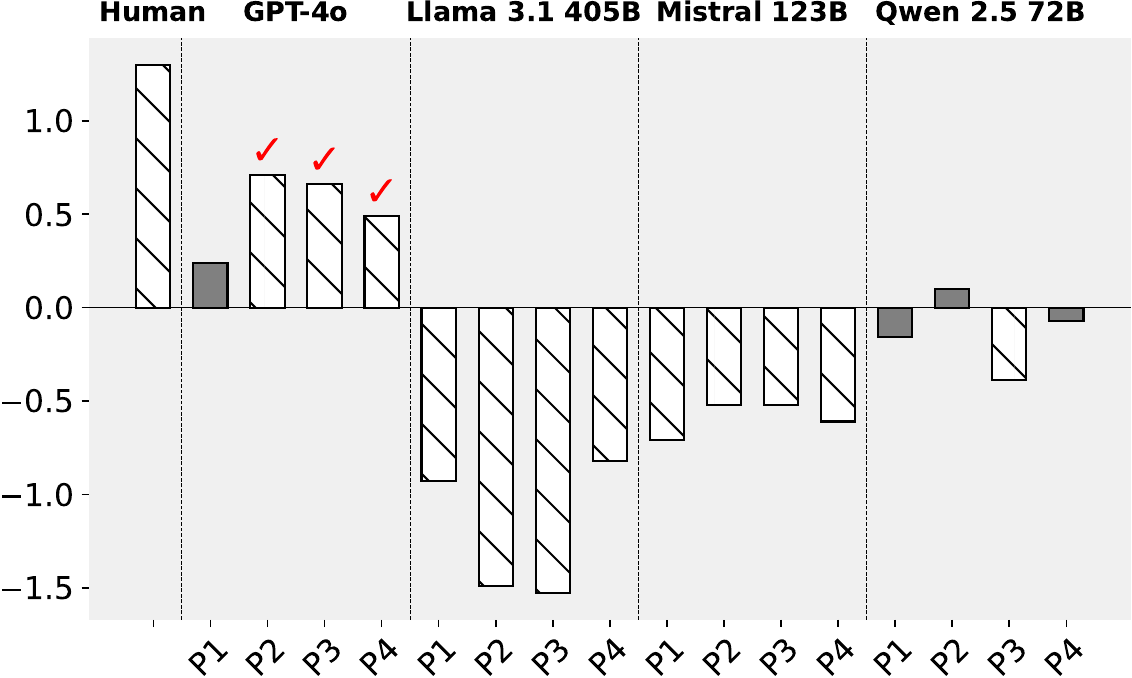}
        \caption{First-person singular pronouns.}
        \label{fig:first-person-singular}
    \end{subfigure}
    \begin{subfigure}{\linewidth}
        \includegraphics[width=0.95\linewidth]{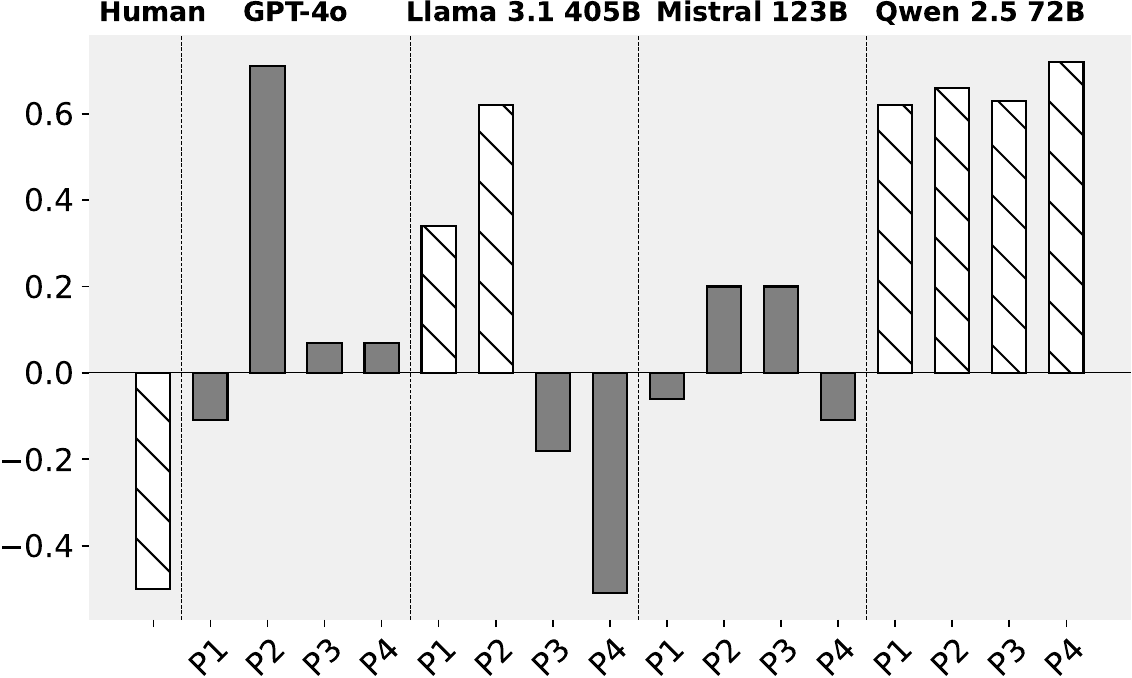}
        \caption{First-person plural pronouns.}
        \label{fig:first-person-plural}
    \end{subfigure}    
    \caption{
    Results for the simple agent using LLMs from each model family versus human.
    P1 to P4 correspond to the prompts presented in \Cref{tab:persona-prompts}.
    Gray bars indicate the result is not statistically significant.
    LLMs barely demonstrate human-like pronoun usage patterns in our experiments.
    }
    \label{fig: simple-agent-advanced-models}
\end{figure}

\begin{figure}[t]
    \centering
    \begin{subfigure}{0.45\linewidth}
        \includegraphics[width=0.95\linewidth]{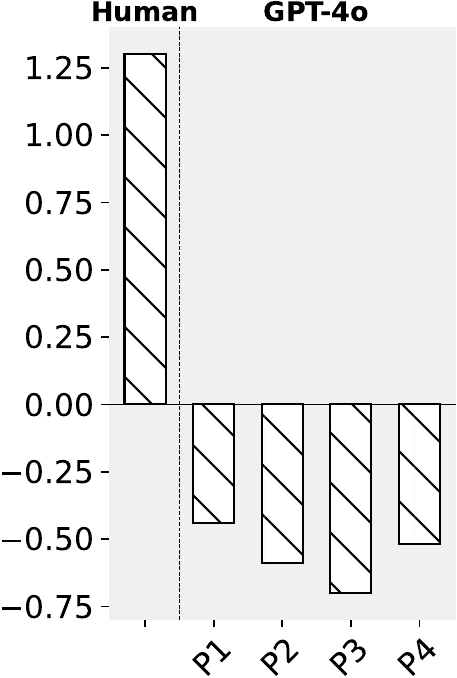}
        \caption{First-person singular.}
        \label{fig:first-person-singular-complex}
    \end{subfigure}
    \begin{subfigure}{0.45\linewidth}
        \includegraphics[width=0.95\linewidth]{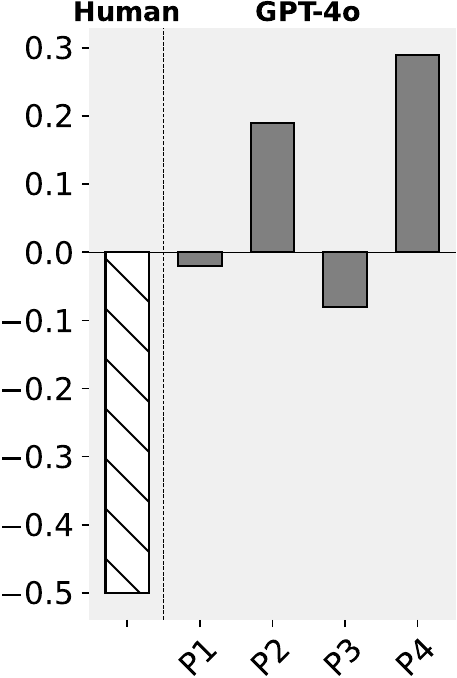}
        \caption{First-person plural.}
        \label{fig:first-person-plural-complex}
    \end{subfigure}    
    \caption{Results for the specialized agent (\Cref{subsec: llm-agent}) based on GPT-4o.
    None of the prompts elicit pronoun usage patterns similar to those of humans.
    In most cases, the observed trends are contrary to human patterns.}
    \label{fig: complex-agents}
\end{figure}

\Cref{fig: simple-agent-advanced-models,fig: complex-agents,fig: simple-agent-llama-family,fig: simple-agent-gpt-family,fig: simple-agent-qwen-family} report the difference of first-person singular and first-person plural pronoun frequencies between non-leaders and leader's scripts.
In \Cref{fig: simple-agent-advanced-models,fig: complex-agents,fig: simple-agent-llama-family,fig: simple-agent-gpt-family,fig: simple-agent-qwen-family}, gray bars indicate that the results are not statistically significant.
We place a red check mark on top of a bar if it demonstrates the same pattern as humans (e.g. in \Cref{fig: simple-agent-advanced-models}, the bars corresponding to GPT-4o with Prompt 2, 3, 4 are positive with statistical significance, same as human).
We report additional results corresponding to all pronouns, and an example of the model's responses when it acts as different roles in \Cref{app-sec: detailed-experimental-results}.

\subsection{Do LLM agents demonstrate human-like pronoun usage patterns?}
\paragraph{Answer: Barely.}
\Cref{fig: simple-agent-advanced-models} presents the results for GPT-4o, Llama 3.1 405B, Mistral 123B, and Qwen 2.5 72B models. 
Only GPT-4o, when prompted with Prompt 2, 3, and 4 (prompts in \Cref{tab:persona-prompts}), exhibits human-like pronoun usage patterns for first-person singular pronouns. 
The other models do not show similar patterns to humans, regardless of the prompt. 
Furthermore, for first-person singular pronouns, Llama 3.1 405B and Mistral 123B display trends opposite to humans. 
For first-person plural pronouns, Llama 3.1 405B and Qwen 2.5 72B show trends contrary to humans.
In \Cref{app-subsec: other-effects}, we try to anonymize names, assign all the agents as males or females, or explicitly prompt the agent to use certain pronouns more often.
However, None of these attempts succeeds in eliciting human-like pronoun usage patterns in LLMs.
Therefore, we conclude that \textit{interactions of agents based on LLMs from each LLM family barely resemble patterns of human interactions.}
From a psychological standpoint, pronoun usage is closely tied to self-identity, social relationships, and group dynamics \citep{gumperz1982language}. 
Humans naturally adjust their use of singular and plural pronouns to convey individuality or solidarity, often without conscious effort. 
Such unconscious behaviors pose significant challenges for LLM agents to replicate in their simulation process.

\subsection{Does the specialized agent demonstrate human-like pronoun usage patterns?}
\paragraph{Answer: No.}
\Cref{fig: complex-agents} presents the results for the specialized agent described in \Cref{subsec: llm-agent} based on GPT-4o. 
We evaluate the effects of the four persona prompts listed in \Cref{tab:persona-prompts}, none of which produce patterns resembling human behavior. 
Interestingly, the simple agent without additional components enables GPT-4o to mimic human pronoun usage patterns with three out of the four prompts (\Cref{fig:first-person-singular}). 
This suggests that \textit{incorporating components like reflection or planning may not bridge the gap but instead deviate pronoun usage patterns further from human norms}.

From a cognitive perspective, this finding aligns with how intuitive processing is involved in human communication. 
Humans often rely on unconscious, automatic processes when engaging in social interactions, effortlessly navigating complex linguistic and social cues without explicit deliberation \citep{bargh1989conditional, bargh2012automaticity}. 
By adding components like reflection or planning to the model, we may introduce a level of explicit reasoning that disrupts the natural flow of language generation, leading to less human-like interactions among the LLM agents.

\begin{figure}[t]
\centering
    \begin{subfigure}{\linewidth}
        \includegraphics[width=0.95\linewidth]{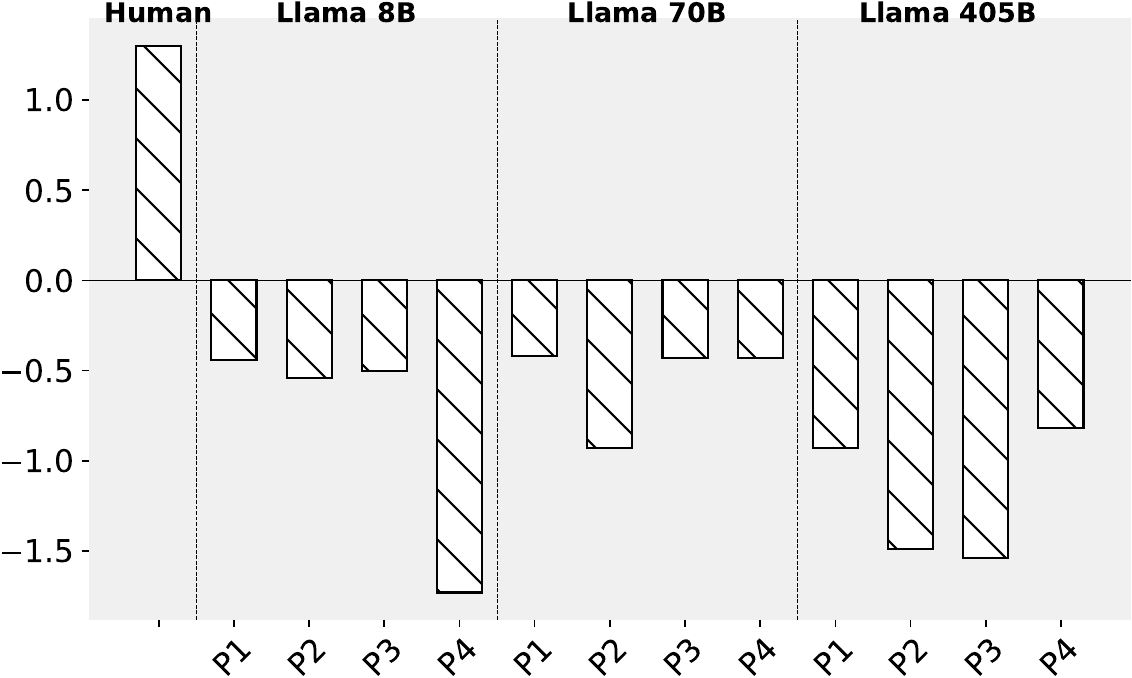}
        \caption{First-person singular pronouns.}
        \label{fig:first-person-singular-llama}
    \end{subfigure}
    \begin{subfigure}{\linewidth}
        \includegraphics[width=0.95\linewidth]{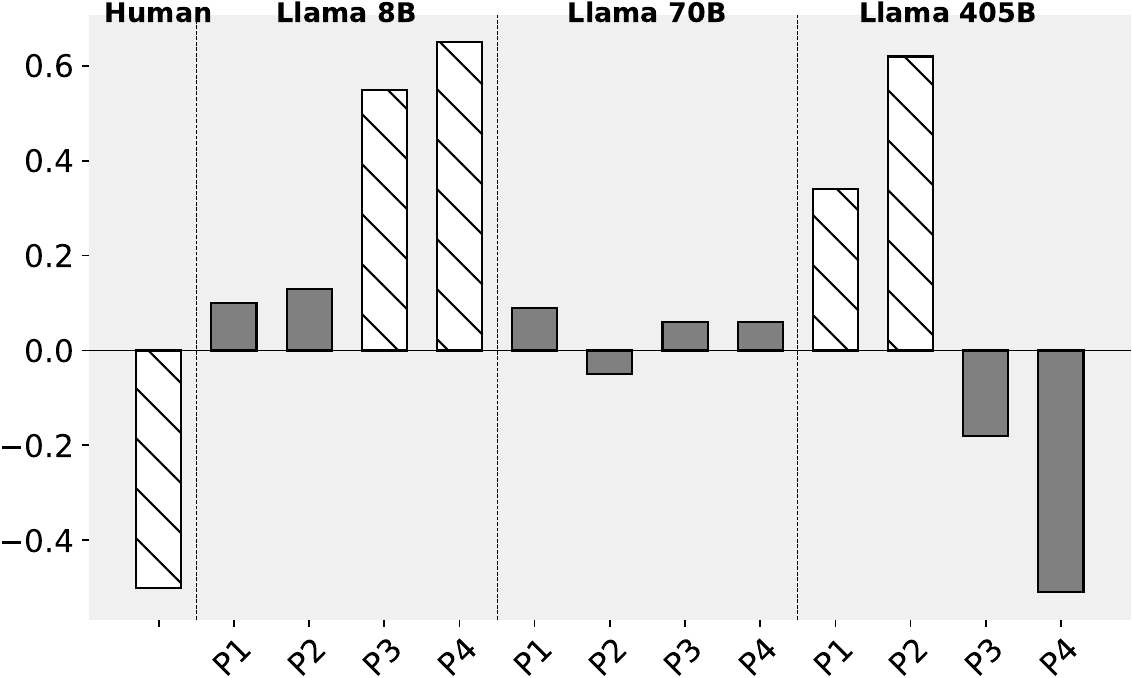}
        \caption{First-person plural pronouns.}
        \label{fig:first-person-plural-llama}
    \end{subfigure}   
    \caption{
    Results for the simple agent based on the Llama models.
    None of the settings elicit human-like behaviors.
    }
    \label{fig: simple-agent-llama-family}
    \vspace{-0.5em}
\end{figure}

\begin{figure}[t]
\centering
    \begin{subfigure}{\linewidth}
        \includegraphics[width=0.95\linewidth]{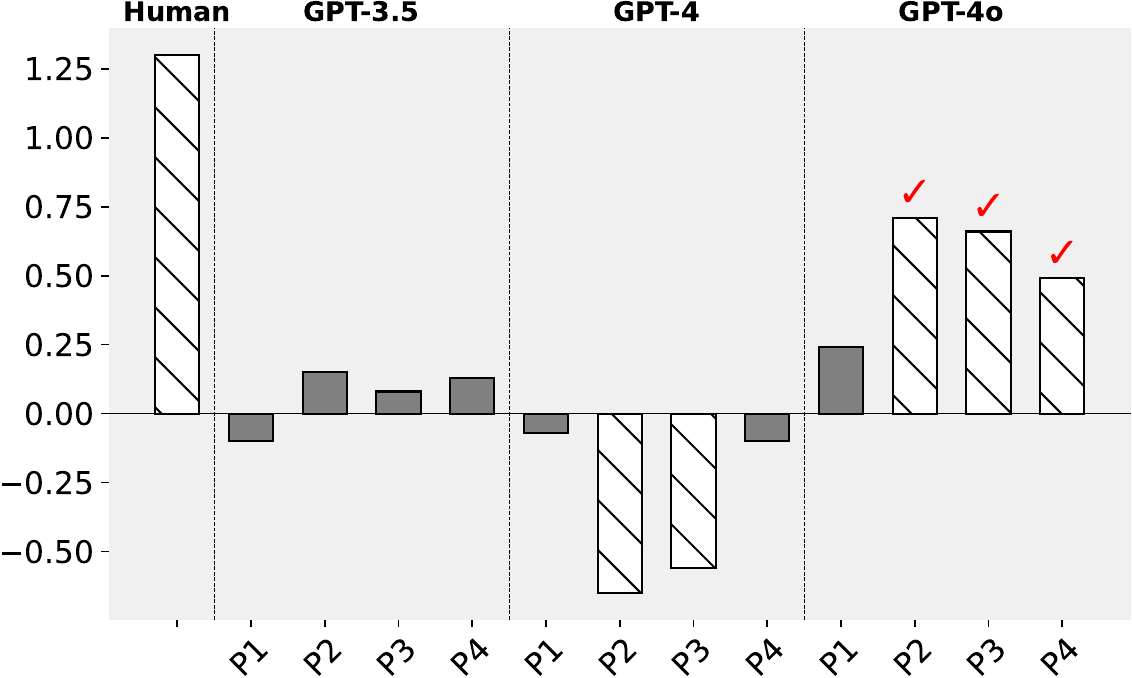}
        \caption{First-person singular pronouns.}
        \label{fig:first-person-singular-gpt}
    \end{subfigure}
    \begin{subfigure}{\linewidth}
        \includegraphics[width=0.95\linewidth]{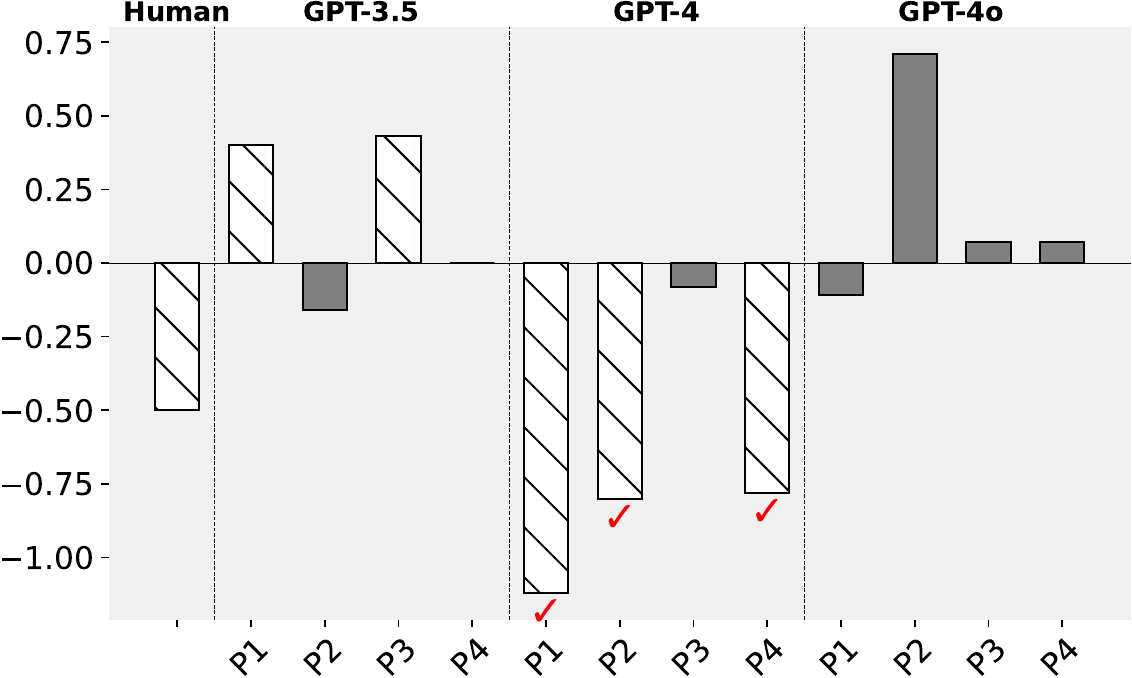}
        \caption{First-person plural pronouns.}
        \label{fig:first-person-plural-gpt}
    \end{subfigure}    
    \caption{
    Results for the simple agent based on GPT models.
    For first-person singular pronouns, agents based on GPT-4o show the same trends as humans for Prompt 2, 3, and 4, while for first-person plural pronouns, agents based on GPT-4 show the same trends as humans for Prompt 1, 2 and 4.
    }
    \label{fig: simple-agent-gpt-family}
    \vspace{-0.5em}
\end{figure}

\begin{figure*}[t]
\centering
    \begin{subfigure}{\linewidth}
        \includegraphics[width=0.95\linewidth]{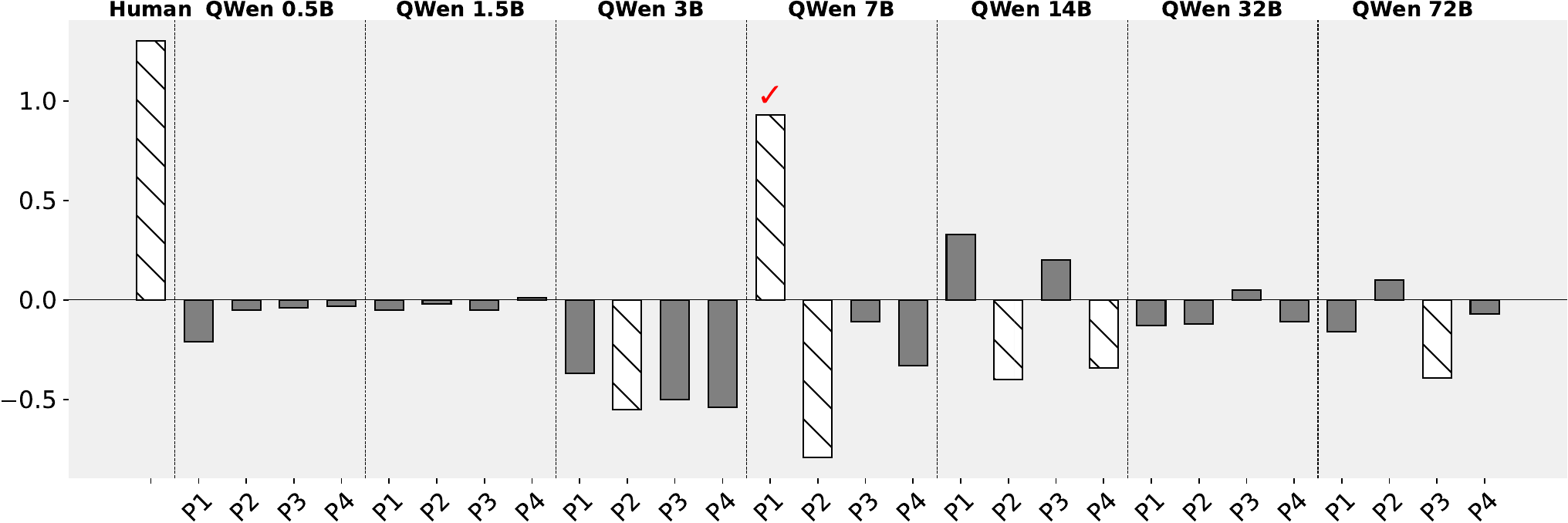}
        \caption{First-person singular pronouns.}
        \label{fig:first-person-singular-qwen}
    \end{subfigure}
    \begin{subfigure}{\linewidth}
        \includegraphics[width=0.95\linewidth]{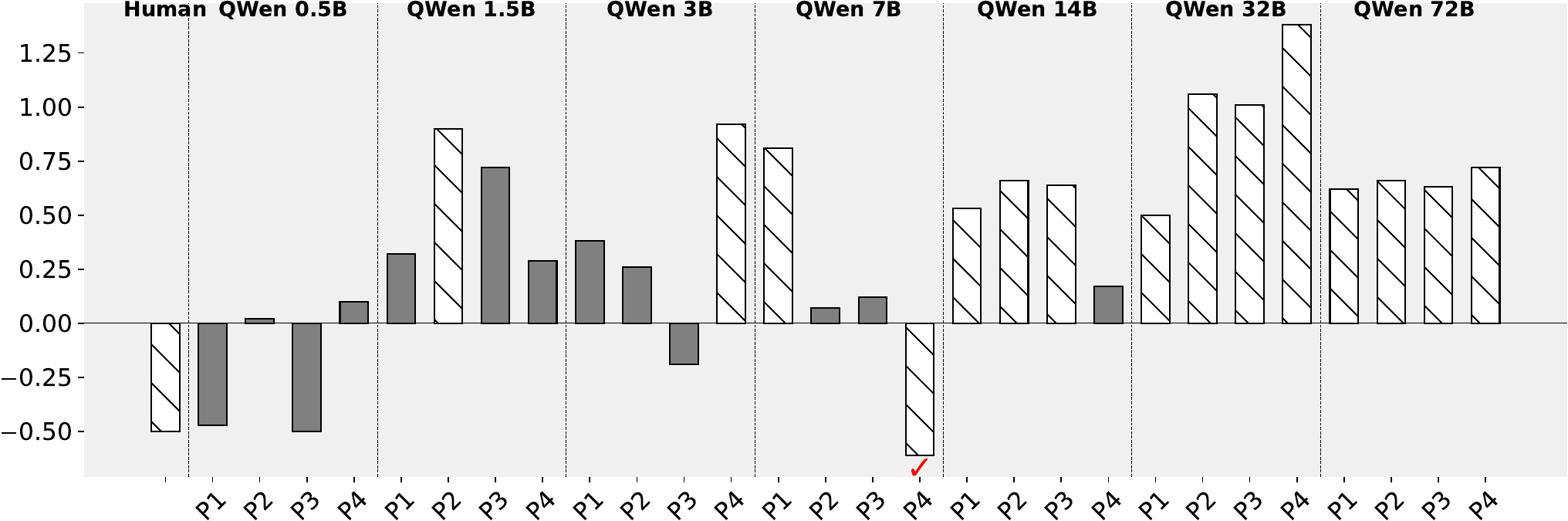}
        \caption{First-person plural pronouns.}
        \label{fig:first-person-plural-qwen}
    \end{subfigure}   
    \caption{
    Results for the simple agent using LLMs from the QWen 2.5 family.
    Only the 7B model with Prompt 1 elicits human-like interaction patterns for first-person singular pronouns, and the 7B model with Prompt 4 for first-person plural pronouns.
    }
    \label{fig: simple-agent-qwen-family}
    \vspace{-0.5em}
\end{figure*}

\subsection{Do LLMs within the same family exhibit consistent pronoun usage patterns?}
\paragraph{Answer: Yes within the same version.}
For the LLMs from the same version, (e.g. Llama 3.1 or QWen 2.5), they show similar trends of pronoun usage patterns across different sizes.
\Cref{fig: simple-agent-llama-family,fig: simple-agent-qwen-family} show the results for LLMs of different sizes from Llama 3.1 and QWen 2.5, respectively.  
In \Cref{fig:first-person-singular-llama}, all the Llama 3.1 models demonstrate a statistically significant trend opposite to the human interaction patterns, regardless of the prompts.
In \Cref{fig:first-person-singular-qwen}, most QWen 2.5 models demonstrate statistically insignificant results in terms of the first-person singular pronouns.
We hypothesize that as models from the same version are trained on similar or the same data, such data similarity leads to the similar interaction patterns regardless of their sizes.

\Cref{fig: simple-agent-gpt-family} shows results for different versions of LLMs from the GPT family.
For first-person singular pronouns, GPT-4o shows similar trends to humans with Prompt 2, 3, and 4, while for first-person plural pronouns, GPT-4 shows similar trends to humans with Prompt 1, 2, and 4.
In contrast, none of the prompts elicit human-like patterns for GPT-3.5.
We hypothesize that as GPT-4 and GPT-4o are more capable than GPT-3.5, they may be better at role-playing and demonstrating human-like pronoun usage patterns.

In addition, \textit{the larger LLM from the same family does not necessarily elicit more human-like pronoun usage patterns}.
For instance, in \Cref{fig:first-person-plural-qwen}, the QWen 72B models demonstrate an opposite pronoun usage pattern compared to humans regardless of the prompts.
In contrast, though in most cases, QWen models fail to elicit human-like pronoun patterns, the QWen 7B model with Prompt 4 elicits human-like patterns for the first-person plural pronouns.


\subsection{Do LLMs ``know'' the pronoun patterns better than what they demonstrate in interactions?}

\begin{table}[t]
    \small
    \centering
    \renewcommand{\arraystretch}{1.3}
    \scalebox{0.95}{
    \begin{tabular}{c|ccccc}
    \toprule
    \multirow{2}{*}{Models} & \multicolumn{2}{c|}{\begin{tabular}[c]{@{}c@{}} First-Person\\	Singular\end{tabular}} & \multicolumn{2}{c|}{\begin{tabular}[c]{@{}c@{}} First-Person\\Plural\end{tabular}} \\
    \cline{2-5}
    & \multicolumn{1}{c|}{kn.?} &  \multicolumn{1}{c|}{dem.?} & \multicolumn{1}{c|}{kn.?} &  \multicolumn{1}{c|}{dem.?} \\
    \hline
    Gold & \ck 6 & \cd 4 & \ck 6 & \cd 4 \\
    \multicolumn{5}{l}{\grayc \textit{GPT}}\\
    3.5 & \ck 0 & \cd 0 & \ck 2 & \cd 0\\
    4  & \ck 6 & \cd 0 & \ck 3 & \cd 3\\
    4o & \ck 5 & \cd 3 & \ck 5 & \cd 0\\
    \multicolumn{5}{l}{\grayc \textit{Llama 3.1 Instruct}}\\
    8B & \ck 0 & \cd 0 & \ck 4 & \cd 0\\
    70B & \ck 6 & \cd 0 & \ck 0 & \cd 0\\
    405B & \ck 1 & \cd 0 & \ck 3 & \cd 0\\
    \multicolumn{5}{l}{\grayc \textit{Qwen 2.5 Instruct}}\\
    0.5B & \ck 0 & \cd 0 & \ck 0 & \cd 0\\
    1.5B & \ck 0 & \cd 0 & \ck 2 & \cd 0\\
    3B & \ck 1 & \cd 0 & \ck 2 & \cd 0 \\
    7B  & \ck 5 & \cd 1 & \ck 5 & \cd 1\\
    14B & \ck 5 & \cd 0 & \ck 6 & \cd 0\\
    32B & \ck 3 & \cd 0 & \ck 0 & \cd 0\\
    72B & \ck 6 & \cd 0 & \ck 3 & \cd 0\\
    \multicolumn{5}{l}{\grayc \textit{Mistral Instruct}}\\
    123B & \ck 3 & \cd 0 & \ck 5 & \cd 0\\
    \bottomrule
    \end{tabular}
    }
    \caption{
    ``Kn.'' denotes with respect to the perturbed prompts, how many times the LLM answers the question correctly.
    ``Dem.'' denotes with respect to the four persona prompts, how many times the agents based on each LLM demonstrate the human-like pronoun pattern during the interaction process.
    The disparity of ``kn.'' and ``dem.'' in LLMs demonstrates that even if the LLM knows the pronoun patterns, they do not necessarily demonstrate them in the interaction process. 
    }
    \label{tab:prompt-and-demonstrate}
    \vspace{-0.5em}
\end{table}

\paragraph{Answer: Yes.}
To understand whether these LLMs ``know'' the underlying social interaction patterns, we employ a variety of prompts based on the following prompt template,

\begin{mybox}{Query Prompt}
For the frequency of the usage for \{\textsc{Pronoun Type}\}, who uses it at a higher frequency in their interactions? Please select one of the three answers: leader, subordinate, equal, please answer in JSON format of \{``answer'': <Your Choice>\}    
\end{mybox}

\noindent Specifically, as the answer orders may influence the results \citep{zheng2023large}, to make the results robust, we permute the answers to prompt each LLM on six different orders for ``leader, subordinate, equal'' (e.g. ``leader, equal, subordinate'', ``subordinate, leader, equal'', etc.).
In \Cref{tab:prompt-and-demonstrate}, we combine our results here with the results from \Cref{sec: results} on whether the simple agents based on each LLM know and demonstrate human-like pronoun usage patterns.

In \Cref{tab:prompt-and-demonstrate}, we find that GPT-4o succeeds five out of six times in selecting the right option for first-person singular pronouns, and succeeds in demonstrating such a pattern in its interaction for three out of the four prompts.
However, for most LLMs, we find that \textit{they may know the human pronoun usage patterns, but do not demonstrate them in their interactions}.
For instance, LLMs such as Llama 70B can correctly choose the option in all six cases for the first-person singular pronoun, but fail to demonstrate such patterns across the four prompts.
\citet{wu-etal-2023-hi} present a similar finding of the disparity between the LLM’s knowledge versus their ability to effectively apply that knowledge in their reasoning process.
Similarly, our findings reveal the disparity between LLMs' knowledge versus demonstrating it in practice.

\section{Discussions}
\label{sec: discussion}



Taking our results together, firstly, \textit{the LLMs' pronoun usage patterns are susceptible to model choices and prompt designs.}
In \Cref{fig: simple-agent-advanced-models}, GPT-4o fails to demonstrate human-like interaction patterns with Prompt 1 even though it succeeds with Prompt 2, 3, and 4.
In \Cref{fig: simple-agent-gpt-family}, though Prompt 1 elicits human-like pronoun usage patterns for the GPT-4 model on first-person plural pronouns, it fails for the GPT-4o model.
Such a model and prompt-susceptible pattern is significantly different from what we expect in humans.
A rational human being would behave as instructed regardless of the wording changes, as long as the core information of playing a \{\textsc{Persona}\} are the same within those instructions.
Such a discrepancy poses concerns about how much we can rely on LLMs for trustworthy human interaction simulations.

Secondly, \textit{for researchers who study social problems through the LLM simulations, they need to be mindful that they are studying LLMs' stochastic group behaviors rather than human behaviors.}
While currently, the community seems to overly focus on the promising ``emergent social behaviors'' and ``social prototyping'' from \citet{park2022social, park2023generative}, we want to emphasize the unpleasant side of the story, that these LLMs are not humans, and the ``emergent social behaviors'' are essentially the LLMs' social behaviors rather than humans'.
In our case, if a researcher is to study the ``emergent pronoun usage'' behaviors of GPT-3.5 in a society of hierarchical structure (the leaders and non-leaders), their conclusion may flip the well-established findings from the psychology community.
This poses concerns to the existing studies where researchers employ LLMs for social simulation, that how authentic and reliable the intermediate process is in the simulation.
Therefore, we advocate for researchers to investigate the underlying interaction mechanisms carefully before they generalize their conclusions to human interactions or human society.

A concurrent study by \citet{park2024generative} demonstrates that, through careful prompt design, LLM agents can replicate participants' responses with 85\% accuracy. 
While their work focuses on simulating individual survey responses, our research emphasizes pronoun usage in LLM interactions. 
Notably, despite their optimization efforts, \citet{park2024generative} only achieve 85\% accuracy on 1,000 individuals, raising questions about the generalizability of their findings to larger populations. 
Furthermore, the underlying reasoning process of LLMs remains opaque, leaving questions about what features LLMs leverage in achieving such results.


\section{Conclusion}
In this paper, we take the first step to study whether LLM agents can demonstrate human-like pronoun usage patterns in the multi-agent interaction process.
By simulating experimental setups from an existing psychology study, we reveal that LLM agents barely demonstrate human-like pronoun patterns, even if the LLM agent may show some understanding of those patterns.
We advocate practitioners scrutinize their simulations and understand the limitations of the LLM-based social simulation.

\section*{Limitations}
We address two primary concerns raised about our work.
First, one might argue that differences in singular or plural pronoun usage are merely formatting issues and do not impact the final simulation outcomes. 
However, we emphasize that if LLMs fail to replicate these subtle but critical markers in human interactions, it raises concerns about their ability to authentically model the underlying human interaction process or the complex social dynamics.
Therefore, we argue that researchers need to be cautious about drawing conclusions from the ``emergent social behaviors'' of LLM agents' interactions.  

Second, some may suggest that better prompt engineering could elicit more human-like pronoun usage.
We want to emphasize that for the scope of this work, we systematically evaluate existing prompts employed by researchers, and they consistently fail to elicit human-like leader pronoun patterns.
If a researcher simply relies on these prompts and LLM agents to simulate human society, they may draw conclusions on pronoun usage patterns that contradict to human societal norms.
In addition, we highlight that when humans act as leaders, they cannot control the frequency of ``we'' or ``us'' consciously and their behaviors are not influenced by explicit ``prompt design''.
Even if optimized prompts can achieve human-like behaviors in LLMs, the optimization process itself underscores the differences between LLMs and human beings.
Finally, such optimization is likely to create a ``rob Peter to pay Paul'' scenario, where improving performance for specific cases results in failures in others, limiting the generalizability of LLMs as robust social simulators.

We try our best to experiment with different LLMs.
However, due to the scope of this study and the limited budget, we cannot test every possible LLM model.
We believe the experiments we conduct in our paper provide strong evidence to our claims.

\section*{Ethics Statement}
In this work, we simulate an existing psychology study by employing LLM agents.
Our results indicate that there is a significant discrepancy in pronoun usage patterns between LLM agents and humans.
We suggest researchers realize the limitations of employing LLM agents to simulate human interactions or human societies, as there are human patterns that cannot be easily and consistently replicated by existing LLMs.

\section*{Acknowledgement}
We thank Yunxiang Zhang, Longju Bai, Zhenjie Sun, Ruiqi He, Hanchen Xia, Angana Borah for their constructive feedback on this paper.

\bibliography{anthology,custom}
\bibliographystyle{acl_natbib}

\newpage
\appendix
\newpage

\section{Additional Experimental Setups}
\label{app-sec: additional-experimental-setups}

\begin{table*}[t]
    \small
    \centering
    \begin{tabular}{p{8em}p{40em}}
    \toprule
    \multicolumn{2}{l}{\grayc Pools} \\
    Male Names & James, John, William, Robert, Michael, David, Richard, Joseph, Charles, Thomas, Daniel, Matthew, Anthony, Mark, Andrew, Joshua, Christopher, Jonathan, Benjamin, Samuel, Henry, Alexander, George, Edward, Nicholas, Luke, Jack, Ethan, Oliver, Jacob \\
    \midrule
    Female Names & Emma, Olivia, Sophia, Ava, Isabella, Mia, Amelia, Harper, Evelyn, Abigail, Emily, Ella, Grace, Lily, Chloe, Scarlett, Victoria, Madison, Aria, Zoey, Nora, Charlotte, Hannah, Lucy, Zoe, Layla, Penelope, Lillian, Natalie, Violet, Stella \\
    \midrule
    Company Names & Company Mars, Company Zelda, Company Moon, Company Jupiter \\
    \midrule
    Ranking Items & 1. Invest in training customer service representatives.

2. Embrace omnichannel support.

3. Collect customer feedback.

4. Use AI and automation tools.

5. Personalize customer interactions.

6. Provide comprehensive FAQs.

7. Monitor and respond to social media. \\
    \bottomrule
    \multicolumn{2}{l}{\grayc Phase I} \\
    Instruction     &  Now your team is tasked with improving the customer service quality of \{COMPANY\}.
Given the following ten ranking items: 

\{RANKING\_ITEMS\}

Now please work on your own and rank these proposed items in order of their positive contribution to the company culture.

Please speak in the format of: \{\{\{NAME\} (\{ROLE\}): <Your Speech>\}\}\\
         \bottomrule
         \multicolumn{2}{l}{\grayc Phase II} \\
         Instruction (Leader) & Now you are discussing in a group.
The topic is recommendation of strategies for service quality improvement by ranking these proposed items in order of their positive contribution to such the company culture for \{company\}. 
The final team decision does not need to be a consensus. 
You as a leader ultimately have the authority to make the final decision, and everyone on the team is clear about it.     
Please speak in the format of: \{\{\{NAME\} (\{ROLE\}): <Your Speech>\}\}\\
Instruction (Non-Leader) & Now you are discussing in a group.
The topic is recommendation of strategies for service quality improvement by ranking these proposed items in order of their positive contribution to such the company culture for \{company\}. 
The final team decision does not need to be a consensus. 
The leader ultimately has the authority to make the final decision, you and everyone on the team are clear about it.       
Please speak in the format of: \{\{\{NAME\} (\{ROLE\}): <Your Speech>\}\}\\
    \bottomrule
    \end{tabular}
    \caption{Pools of items that we use to populate the masked sections in instructions, along with the instructions that we use in the two experimental phases. 
    The setup involves two roles: leader and subordinate.
    }
    \label{tab:instructions}
\end{table*}

\paragraph{Simulation Scenarios.}
\Cref{tab:instructions} presents the pools for male names, female names, company names, and ranking items where we draw to populate the masked sections in the instructions, together with the instructions that we use.

In our simulation, the group is charged with completing a group decision-making task, where they act as consulting groups to improve the customer service quality of a fictitious company.
Specifically, the agents need to rank items (e.g. ``Invest in training customer service representatives'', ``Personalize customer interactions'', etc.) in order of their positive contribution.
All the agents are aware that the team decision does not need to be a consensus, and the leader ultimately has the authority to make the final decision.

\paragraph{Model Hyperparameters.} For all the models, we employ a temperature of 0.7 to ensure diversity in their generations. 

\paragraph{Statistical Significance Test.}
We collect the pronoun usage frequencies for the leader and non-leader agents during their interactions.
Since we repeat our experiments 41 times consistent with \citet{burris2009playing, kacewicz2014pronoun}, we have 41 frequency values for leaders and non-leaders respectively for one type of pronoun.
We then conduct the statistical significance T-test to calculate the two samples of frequencies to get the p values.

\section{Additional Experimental Results}
\label{app-sec: detailed-experimental-results}

\begin{table*}[t]
    \centering
    \begin{subtable}{0.45\textwidth}
        \small
        \centering
        \renewcommand{\arraystretch}{1.3}
        \setlength\tabcolsep{4pt}
        \resizebox{\linewidth}{!}{
        \begin{tabular}{lcccc}
        \toprule
          &  Examples & Leader & Nonleaders & \textit{p}\\
          \midrule
        Word Counts &  - & 1076.90 & 645.00 & .00\\
            \midrule
        All pronouns & -  & 19.80 & 20.80 & .01\\
        \quad Personal pronouns & I, we, she & 10.10 & 10.60 & .06 \\
        \quad \quad First-person singular & I, me & 4.30 & 5.60 & .00\\
        \quad \quad First-person plural & We, us & 2.10 & 1.60 & .00\\
        \quad \quad Second-person & You, your & 2.70 & 2.50 & .20\\
        \quad \quad Third-person singular & She, he, her & 0.07 & 0.09 & .50\\
        \quad \quad Third-person plural & They, them & 0.83 & 0.85 & .84\\
        \quad Impersonal pronouns & It, that & 9.70 & 10.10 & .06\\   
        \bottomrule
        \end{tabular}
        }
        \caption{Human \citep{kacewicz2014pronoun}.}
        \label{tab:human}
    \end{subtable}~
    \begin{subtable}{0.45\textwidth}
        \small
        \centering
        \renewcommand{\arraystretch}{1.3}
        \setlength\tabcolsep{4pt}
        \resizebox{\linewidth}{!}{
        \begin{tabular}{lcccc}
        \toprule
          &  Examples & Leader & Nonleaders & \textit{p}\\
          \midrule
        Word Counts &  - & 2242.63 & 1946.86 & .21\\
            \midrule
        All pronouns & -  & 6.80 & 6.89 & .98 \\
        \quad Personal pronouns & I, we, she & 5.99 & 5.96 & .73\\
        \quad \quad First-person singular & I, me &  1.84 & 2.08 &.08\\
        \quad \quad First-person plural & We, us & 2.54 & 2.43 & .73\\
        \quad \quad Second-person & You, your & 0.98 & 0.75 & .04 \\
        \quad \quad Third-person singular & She, he, her & 0.03 & 0.03 & .78\\
        \quad \quad Third-person plural & They, them & 0.61 & 0.67 & .61\\
        \quad Impersonal pronouns & It, that & 0.81 & 0.93 & .56\\   
        \bottomrule
        \end{tabular}
        }
        \caption{GPT-4o, prompt 1.}
        \label{tab:gpt-4o-prompt-1}
    \end{subtable}

    \vspace{1em}
    \begin{subtable}{0.45\textwidth}
        \small
        \centering
        \renewcommand{\arraystretch}{1.3}
        \setlength\tabcolsep{4pt}
        \resizebox{\linewidth}{!}{
        \begin{tabular}{lcccc}
        \toprule
          &  Examples & Leader & Nonleaders & \textit{p}\\
          \midrule
        Word Counts & - &  1270.44 &  1021.36 & .07 \\
    All pronouns & - &  6.35 &  7.32 & .01 \\
    \quad Personal pronouns & - &  5.84 &  6.36 & .04 \\
    \quad \quad First-person singular & - &  1.11 &  1.82 & .00 \\
    \quad \quad First-person plural & - &  3.09 &  2.95 & .41\\
    \quad \quad Second-person & - &  1.14 &  0.81 & .02\\
    \quad \quad Third-person singular & - &  0.06 &  0.06 & .12\\
    \quad \quad Third-person plural & - &  0.44 &  0.72 & .00 \\
    \quad Impersonal pronouns & - &  0.65 &  0.96 & .00 \\
        \bottomrule
        \end{tabular}
        }
        \caption{GPT-4o, prompt 2.}
        \label{tab:gpt-4o-prompt-2}
    \end{subtable}~
    \begin{subtable}{0.45\textwidth}
    \small
    \centering
    \renewcommand{\arraystretch}{1.3}
    \setlength\tabcolsep{4pt}
    \resizebox{\linewidth}{!}{
    \begin{tabular}{lcccc}
    \toprule
      &  Examples & Leader & Nonleaders & \textit{p}\\
      \midrule
    Word Counts & - &  1609.46 &  1166.73 &  .01 \\
All pronouns & - &  6.32 &  6.98 &  .00 \\
\quad Personal pronouns & - &  5.57 &  6.11 &  .00 \\
\quad \quad First-person singular & - &  1.40 &  2.06 &  .00\\
\quad \quad First-person plural & - &  2.50 &  2.57 &  .60\\
\quad \quad Second-person & - &  1.22 &  0.84 &  .00\\
\quad \quad Third-person singular & - &  0.06 &  0.03 &  .02\\
\quad \quad Third-person plural & - &  0.38 &  0.61 &  .00\\
\quad Impersonal pronouns & - &  0.75 &  0.87 &  .17 \\
    \bottomrule
    \end{tabular}
    }
    \caption{GPT-4o, prompt 3.}
    \label{tab:gpt-4o-prompt-3}
\end{subtable}

\vspace{1em}
\begin{subtable}{0.45\textwidth}
    \small
    \centering
    \renewcommand{\arraystretch}{1.3}
    \setlength\tabcolsep{4pt}
    \resizebox{\linewidth}{!}{
    \begin{tabular}{lcccc}
    \toprule
      &  Examples & Leader & Nonleaders & \textit{p}\\
      \midrule
    Word Counts & - &  1578.83 &  1371.24 &  .30 \\
All pronouns & - &  5.98 &  6.45 &  .06 \\
\quad Personal pronouns & - &  5.15 &  5.61 &  .04 \\
\quad \quad First-person singular & - &  1.42 &  1.91 &  .00\\
\quad \quad First-person plural & - &  2.20 &  2.27 &  .63\\
\quad \quad Second-person & - &  1.07 &  0.80 &  .00\\
\quad \quad Third-person singular & - &  0.01 &  0.01 &  .83\\
\quad \quad Third-person plural & - &  0.46 &  0.62 &  .03\\
\quad Impersonal pronouns & - &  0.82 &  0.85 &  .81 \\
    \bottomrule
    \end{tabular}
    }
    \caption{GPT-4o, prompt 4.}
    \label{tab:gpt-4o-prompt-4}
\end{subtable}
\caption{Complete results of employing GPT-4o as the agent.
Since we primarily care about whether there is a statistically significant difference in the pronoun usage between leaders and non-leaders, we decide to use the difference rather than the absolute frequency as our evaluation metrics in \Cref{eq: eval-calc}.}
\end{table*}

\subsection{Absolute Pronoun Usage Results.}
\Cref{tab:gpt-4o-prompt-4} shows the results we get for each pronoun usage by employing GPT-4o as the agent.
We structure each table in the same format as \Cref{tab:human} from \citet{kacewicz2014pronoun}.
We note that we report the frequency in terms of percentage ($f\in[0, 100]$) following \citet{kacewicz2014pronoun}.

\begin{table*}[t]
    \small
    \centering
    \renewcommand{\arraystretch}{1.3}
    \setlength\tabcolsep{4pt}
    \scalebox{1}{
    \begin{tabular}{cccccccccc}
    \toprule
        & & \begin{tabular}[c]{@{}c@{}}\normalsize All\\\normalsize Pronouns\end{tabular} & \begin{tabular}[c]{@{}c@{}}Personal\\Pronouns\end{tabular} & \begin{tabular}[c]{@{}c@{}}\scriptsize	First-Person\\\scriptsize	Singular\end{tabular} & \begin{tabular}[c]{@{}c@{}}\scriptsize	First-Person\\\scriptsize	Plural\end{tabular} & \begin{tabular}[c]{@{}c@{}}\scriptsize	Second-\\\scriptsize	Person\end{tabular} & \begin{tabular}[c]{@{}c@{}}\scriptsize Third-Person\\\scriptsize Singular\end{tabular} & \begin{tabular}[c]{@{}c@{}}\scriptsize Third-Person\\\scriptsize Plural\end{tabular} & \begin{tabular}[c]{@{}c@{}}Impersonal\\Pronouns\end{tabular} \\
        \midrule
     & Human (\citeyear{kacewicz2014pronoun}) & $1.00$ & $\g0.50$ & $1.30$ & \hspace{-0.77em}$-0.50$ & \hspace{-0.77em}$\g-0.20$ & $\g0.02$ & $\g0.02$ & $\g0.40$ \\
    \multicolumn{10}{l}{\grayc \textit{GPT-4o}}\\
     & Prompt 1 & $\g0.09$ & \hspace{-0.77em}$\g-0.03$ & $\g0.24$ & \hspace{-0.77em}$\g-0.11$ & \hspace{-0.77em}$-0.23$ & $\g0.00$ & $\g0.06$ & $\g0.12$ \\
    & Prompt 2 & $0.97$ & $0.62$ & $0.71$ & $\g0.71$ & \hspace{-0.77em}$-0.33$ & $\g0.00$ & $0.28$ & $0.31$\\
    & Prompt 3 & $0.66$ & $0.54$ & $0.66$ & $\g0.07$ & \hspace{-0.77em}$-0.38$ & \hspace{-0.77em}$-0.03$ & $0.23$ & $0.12$ \\
    & Prompt 4 & $\g0.47$ & $0.46$ & $0.49$ & $\g0.07$ &  \hspace{-0.77em}$-0.27$ & $\g0.00$ & $0.18$ & $\g0.03$ \\
    \multicolumn{10}{l}{\grayc \textit{Llama 3.1 405B Instruct}}\\
    & Prompt 1 & $\g 1.82$ & $\g 0.16$ & \hspace{-0.77em}$-0.93$ & $ 0.34$ & $ 0.80$ & \hspace{-0.77em}$\g-0.01$ & \hspace{-0.77em}$\g-0.03$ & \hspace{-0.77em}$\g-0.04$ \\
    & Prompt 2 & $\g 1.94$ & $\g 0.16$ & \hspace{-0.77em}$-1.49$ & $ 0.62$ & $ 0.82$ & \hspace{-0.77em}$\g-0.09$ & $ 0.30$ & \hspace{-0.77em}$\g-0.13$ \\
    & Prompt 3 & $\g 2.58$ & $\g 0.35$ & \hspace{-0.77em}$-1.53$ & \hspace{-0.77em}$\g-0.18$ & $ 2.41$ & \hspace{-0.77em}$\g-0.09$ & \hspace{-0.77em}$\g-0.26$ & \hspace{-0.77em}$\g-0.29$\\
    & Prompt 4 & $ 4.78$ & \hspace{-0.77em}$\g-0.40$ & \hspace{-0.77em}$-0.82$ & \hspace{-0.77em}$\g-0.51$ & $ 1.04$ & $\g 0.01$ & \hspace{-0.77em}$\g-0.12$ & \hspace{-0.77em}$\g-0.40$ \\
    \multicolumn{10}{l}{\grayc \textit{Mistral 128B Large Instruct}}\\
    & Prompt 1 & $\g 0.46$ & \hspace{-0.77em}$\g-0.09$ & \hspace{-0.77em}$-0.71$ & \hspace{-0.77em}$\g-0.06$ & $ 0.83$ & $ 0.04$ & \hspace{-0.77em}$\g-0.19$ & \hspace{-0.77em}$-0.21$ \\
    & Prompt 2 & $\g 1.09$ & $\g 0.35$ & \hspace{-0.77em}$-0.52$ & $\g 0.20$ & $ 0.60$ & $ 0.02$ & $\g 0.06$ & \hspace{-0.77em}$-0.49$ \\
    & Prompt 3 & $\g 1.09$ & $\g 0.35$ & \hspace{-0.77em}$-0.52$ & $\g 0.20$ & $ 0.60$ & $ 0.02$ & $\g 0.06$ & \hspace{-0.77em}$-0.49$ \\
    & Prompt 4 & $\g 0.88$ & \hspace{-0.77em}$\g-0.39$ & \hspace{-0.77em}$-0.61$ & \hspace{-0.77em}$\g-0.11$ & $ 0.45$ & $ 0.02$ & \hspace{-0.77em}$\g-0.14$ & \hspace{-0.77em}$-0.22$\\
    \multicolumn{10}{l}{\grayc \textit{Qwen 2.5 72B Instruct}}\\
    & Prompt 1 & \hspace{-0.77em}$\g-0.09$ & $ 0.83$ & \hspace{-0.77em}$\g-0.16$ & $ 0.62$ & $ 0.56$ & \hspace{-0.77em}$\g-0.03$ & \hspace{-0.77em}$-0.15$ & \hspace{-0.77em}$-0.43$ \\
    & Prompt 2 & $ 0.21$ & $ 2.09$ & $\g 0.10$ & $ 0.66$ & $ 1.84$ & \hspace{-0.77em}$\g-0.02$ & \hspace{-0.77em}$-0.48$ & \hspace{-0.77em}$-0.66$ \\
    & Prompt 3 & $\g 0.01$ & $ 0.87$ & \hspace{-0.77em}$-0.39$ & $ 0.63$ & $ 0.86$ & $\g 0.00$ & \hspace{-0.77em}$-0.22$ & \hspace{-0.77em}$-0.40$ \\
    & Prompt 4 & $\hspace{-0.77em}-0.27$ & $ 1.06$ & $\g\hspace{-0.77em}-0.07$ & $ 0.72$ & $ 0.47$ & $\g\hspace{-0.77em}-0.02$ & $\g\hspace{-0.77em}-0.03$ & $\g\hspace{-0.77em}-0.20$ \\
\midrule
\multicolumn{10}{l}{\grayc 	\textit{GPT-4o Framework Agent}}\\
 & Prompt 1 & $\g 0.56$ & $\g\hspace{-0.77em}-0.11$ & $\hspace{-0.77em}-0.44$ & $\g\hspace{-0.77em}-0.02$ & $ 0.41$ & $\g\hspace{-0.77em}-0.00$ & $\g\hspace{-0.77em}-0.05$ & $\g\hspace{-0.77em}-0.12$ \\
 & Prompt 2 & $\g 2.11$ & $\g\hspace{-0.77em}-0.12$ & $\hspace{-0.77em}-0.59$ & $\g 0.19$ & $ 0.35$ & $ 0.04$ & $\g\hspace{-0.77em}-0.10$ & $\hspace{-0.77em}-0.46$ \\
 & Prompt 3 & $ 0.45$ & $\g\hspace{-0.77em}-0.20$ & $\hspace{-0.77em}-0.70$ & $\g\hspace{-0.77em}-0.08$ & $ 0.77$ & $\g 0.02$ & $\hspace{-0.77em}-0.21$ & $\hspace{-0.77em}-0.26$ \\
 & Prompt 4 & $\g 0.09$ & $\g 0.37$ & $\hspace{-0.77em}-0.52$ & $\g 0.29$ & $ 0.88$ & $ 0.05$ & $\hspace{-0.77em}-0.34$ & $\hspace{-0.77em}-0.17$ \\
    \bottomrule
    \end{tabular}
    }
    \caption{
    Results corresponding to \Cref{fig: simple-agent-advanced-models} and \Cref{fig: complex-agents}.
    We report the difference in the average pronoun frequencies between non-leaders and leader's scripts, and put the number in gray if there is no statistical significance, i.e., the $p$-value is smaller than 0.05.}
    \label{tab:aggregated}
\end{table*}

\subsection{Complete Results for Pronoun Usage Difference}

\begin{table*}[t]
    \small
    \centering
    \renewcommand{\arraystretch}{1.3}
    \setlength\tabcolsep{4pt}
    \scalebox{1}{
    \begin{tabular}{cccccccccc}
    \toprule
        & & \begin{tabular}[c]{@{}c@{}}\normalsize All\\\normalsize Pronouns\end{tabular} & \begin{tabular}[c]{@{}c@{}}Personal\\Pronouns\end{tabular} & \begin{tabular}[c]{@{}c@{}}\scriptsize	First-Person\\\scriptsize	Singular\end{tabular} & \begin{tabular}[c]{@{}c@{}}\scriptsize	First-Person\\\scriptsize	Plural\end{tabular} & \begin{tabular}[c]{@{}c@{}}\scriptsize	Second-\\\scriptsize	Person\end{tabular} & \begin{tabular}[c]{@{}c@{}}\scriptsize Third-Person\\\scriptsize Singular\end{tabular} & \begin{tabular}[c]{@{}c@{}}\scriptsize Third-Person\\\scriptsize Plural\end{tabular} & \begin{tabular}[c]{@{}c@{}}Impersonal\\Pronouns\end{tabular} \\
        \midrule
     & Human (\citeyear{kacewicz2014pronoun}) & $1.00$ & $\g0.50$ & $1.30$ & \hspace{-0.77em}$-0.50$ & \hspace{-0.77em}$\g-0.20$ & $\g0.02$ & $\g0.02$ & $\g0.40$ \\
     \multicolumn{10}{l}{\grayc \textit{GPT-3.5}}\\
     & Prompt 1 & \hspace{-0.77em}$\g-0.49$ & $ 0.52$ & \hspace{-0.77em}$\g-0.10$ & $ 0.40$ & $ 0.19$ & \hspace{-0.77em}$\g-0.02$ & $\g 0.05$ & \hspace{-0.77em}$\g-0.03$ \\
     & Prompt 2 & $\g 0.19$ & $\g 0.07$ & $\g 0.15$ & \hspace{-0.77em}$\g-0.16$ & $\g 0.18$ & \hspace{-0.77em}$\g-0.04$ & \hspace{-0.77em}$\g-0.06$ & \hspace{-0.77em}$\g-0.09$ \\
     & Prompt 3 & \hspace{-0.77em}$\g-0.19$ & $ 0.59$ & $\g 0.08$ & $ 0.43$ & $ 0.27$ & \hspace{-0.77em}$\g-0.03$ & \hspace{-0.77em}$\g-0.17$ & \hspace{-0.77em}$\g-0.15$ \\
     & Prompt 4 & \hspace{-0.77em}$-0.35$ & $ 0.44$ & $\g 0.13$ & \hspace{-0.77em}$\g-0.00$ & $ 0.55$ & \hspace{-0.77em}$\g-0.01$ & \hspace{-0.77em}$-0.23$ & $\g 0.05$ \\
     \multicolumn{10}{l}{\grayc \textit{GPT-4}}\\
    & Prompt 1 & $ 4.44$ & \hspace{-0.77em}$\g-0.39$ & \hspace{-0.77em}$\g-0.07$ & \hspace{-0.77em}$-1.12$ & $ 1.14$ & $ 0.01$ & \hspace{-0.77em}$-0.34$ & \hspace{-0.77em}$-0.65$ \\
    & Prompt 2 & $\g 1.57$ & $ 0.51$ & \hspace{-0.77em}$-0.65$ & \hspace{-0.77em}$-0.80$ & $ 2.04$ & $\g 0.06$ & \hspace{-0.77em}$\g-0.15$ & \hspace{-0.77em}$-0.44$ \\
    & Prompt 3 & $\g 0.83$ & $\g 0.20$ & \hspace{-0.77em}$-0.56$ & \hspace{-0.77em}$\g-0.08$ & $ 1.02$ & $ 0.02$ & \hspace{-0.77em}$-0.19$ & \hspace{-0.77em}$-0.24$ \\
    & Prompt 4 & $\g 2.86$ & \hspace{-0.77em}$\g-0.04$ & \hspace{-0.77em}$\g-0.10$ & \hspace{-0.77em}$-0.78$ & $ 0.83$ & \hspace{-0.77em}$-0.01$ & $\g 0.02$ & \hspace{-0.77em}$\g-0.46$ \\
    \multicolumn{10}{l}{\grayc \textit{GPT-4o}}\\
     & Prompt 1 & $\g0.09$ & \hspace{-0.77em}$\g-0.03$ & $\g0.24$ & \hspace{-0.77em}$\g-0.11$ & \hspace{-0.77em}$-0.23$ & $\g0.00$ & $\g0.06$ & $\g0.12$ \\
    & Prompt 2 & $0.97$ & $0.62$ & $0.71$ & $\g0.71$ & \hspace{-0.77em}$-0.33$ & $\g0.00$ & $0.28$ & $0.31$\\
    & Prompt 3 & $0.66$ & $0.54$ & $0.66$ & $\g0.07$ & \hspace{-0.77em}$-0.38$ & \hspace{-0.77em}$-0.03$ & $0.23$ & $0.12$ \\
    & Prompt 4 & $\g0.47$ & $0.46$ & $0.49$ & $\g0.07$ &  \hspace{-0.77em}$-0.27$ & $\g0.00$ & $0.18$ & $\g0.03$ \\
    \bottomrule
    \end{tabular}
    }
    \caption{
    Results for LLM agents based on GPT models corresponding to \Cref{fig: simple-agent-gpt-family}.}
    \label{tab:gpt-family-studies}
\end{table*}

\begin{table*}[t]
    \small
    \centering
    \renewcommand{\arraystretch}{1.3}
    \setlength\tabcolsep{4pt}
    \scalebox{1}{
    \begin{tabular}{cccccccccc}
    \toprule
        & & \begin{tabular}[c]{@{}c@{}}\normalsize All\\\normalsize Pronouns\end{tabular} & \begin{tabular}[c]{@{}c@{}}Personal\\Pronouns\end{tabular} & \begin{tabular}[c]{@{}c@{}}\scriptsize	First-Person\\\scriptsize	Singular\end{tabular} & \begin{tabular}[c]{@{}c@{}}\scriptsize	First-Person\\\scriptsize	Plural\end{tabular} & \begin{tabular}[c]{@{}c@{}}\scriptsize	Second-\\\scriptsize	Person\end{tabular} & \begin{tabular}[c]{@{}c@{}}\scriptsize Third-Person\\\scriptsize Singular\end{tabular} & \begin{tabular}[c]{@{}c@{}}\scriptsize Third-Person\\\scriptsize Plural\end{tabular} & \begin{tabular}[c]{@{}c@{}}Impersonal\\Pronouns\end{tabular} \\
        \midrule
     & Human (\citeyear{kacewicz2014pronoun}) & $1.00$ & $\g0.50$ & $1.30$ & \hspace{-0.77em}$-0.50$ & \hspace{-0.77em}$\g-0.20$ & $\g0.02$ & $\g0.02$ & $\g0.40$ \\
      \multicolumn{10}{l}{\grayc \textit{Llama 3.1 8B Instruct}}\\
      & Prompt 1 & $\g 0.81$ & $\g 0.29$ & \hspace{-0.77em}$-0.44$ & $\g 0.10$ & $ 0.56$ & $\g 0.01$ & $\g 0.05$ & \hspace{-0.77em}$\g-0.24$ \\
      & Prompt 2 & $ 0.69$ & \hspace{-0.77em}$\g-0.18$ & \hspace{-0.77em}$-0.54$ & $\g 0.13$ & $ 0.25$ & $ 0.03$ & \hspace{-0.77em}$\g-0.05$ & \hspace{-0.77em}$-0.34$ \\
      & Prompt 3 & $\g 0.81$ & $\g 0.54$ & \hspace{-0.77em}$-0.50$ & $ 0.55$ & $ 0.66$ & $ 0.02$ & \hspace{-0.77em}$\g-0.18$ & \hspace{-0.77em}$-0.36$ \\
      & Prompt 4 & $ 2.63$ & \hspace{-0.77em}$\g-0.49$ & \hspace{-0.77em}$-1.73$ & $ 0.65$ & $ 0.43$ & $\g 0.03$ & $\g 0.14$ & \hspace{-0.77em}$-0.52$ \\
     \multicolumn{10}{l}{\grayc \textit{Llama 3.1 70B Instruct}}\\
    & Prompt 1 & $\g 0.64$ & \hspace{-0.77em}$\g-0.17$ & \hspace{-0.77em}$-0.42$ & $\g 0.09$ & $\g 0.24$ & \hspace{-0.77em}$\g-0.00$ & \hspace{-0.77em}$\g-0.09$ & \hspace{-0.77em}$\g-0.10$ \\
    & Prompt 2 & $\g 1.44$ & \hspace{-0.77em}$\g-0.42$ & \hspace{-0.77em}$-0.93$ & \hspace{-0.77em}$\g-0.05$ & $ 0.36$ & $\g 0.00$ & $ 0.19$ & \hspace{-0.77em}$\g-0.18$ \\
    & Prompt 2 & \hspace{-0.77em}$\g-0.13$ & $\g 0.40$ & \hspace{-0.77em}$-0.43$ & $\g 0.06$ & $ 0.74$ & $\g 0.08$ & \hspace{-0.77em}$\g-0.03$ & \hspace{-0.77em}$\g-0.06$ \\
    & Prompt 3 & \hspace{-0.77em}$\g-0.13$ & $\g 0.40$ & \hspace{-0.77em}$-0.43$ & $\g 0.06$ & $ 0.74$ & $\g 0.08$ & \hspace{-0.77em}$\g-0.03$ & \hspace{-0.77em}$\g-0.06$ \\
    \multicolumn{10}{l}{\grayc \textit{Llama 3.1 405B Instruct}}\\
    & Prompt 1 & $\g 1.82$ & $\g 0.16$ & \hspace{-0.77em}$-0.93$ & $ 0.34$ & $ 0.80$ & \hspace{-0.77em}$\g-0.01$ & \hspace{-0.77em}$\g-0.03$ & \hspace{-0.77em}$\g-0.04$ \\
    & Prompt 2 & $\g 1.94$ & $\g 0.16$ & \hspace{-0.77em}$-1.49$ & $ 0.62$ & $ 0.82$ & \hspace{-0.77em}$\g-0.09$ & $ 0.30$ & \hspace{-0.77em}$\g-0.13$ \\
    & Prompt 3 & $\g 2.58$ & $\g 0.35$ & \hspace{-0.77em}$-1.53$ & \hspace{-0.77em}$\g-0.18$ & $ 2.41$ & \hspace{-0.77em}$\g-0.09$ & \hspace{-0.77em}$\g-0.26$ & \hspace{-0.77em}$\g-0.29$\\
    & Prompt 4 & $ 4.78$ & \hspace{-0.77em}$\g-0.40$ & \hspace{-0.77em}$-0.82$ & \hspace{-0.77em}$\g-0.51$ & $ 1.04$ & $\g 0.01$ & \hspace{-0.77em}$\g-0.12$ & \hspace{-0.77em}$\g-0.40$ \\
    \bottomrule
    \end{tabular}
    }
    \caption{
    Results for LLM agents based on Llama models corresponding to \Cref{fig: simple-agent-llama-family}.}
    \label{tab:llama-family-studies}
\end{table*}

\begin{table*}[t]
    \small
    \centering
    \renewcommand{\arraystretch}{1.3}
    \setlength\tabcolsep{4pt}
    \scalebox{1}{
    \begin{tabular}{cccccccccc}
    \toprule
        & & \begin{tabular}[c]{@{}c@{}}\normalsize All\\\normalsize Pronouns\end{tabular} & \begin{tabular}[c]{@{}c@{}}Personal\\Pronouns\end{tabular} & \begin{tabular}[c]{@{}c@{}}\scriptsize	First-Person\\\scriptsize	Singular\end{tabular} & \begin{tabular}[c]{@{}c@{}}\scriptsize	First-Person\\\scriptsize	Plural\end{tabular} & \begin{tabular}[c]{@{}c@{}}\scriptsize	Second-\\\scriptsize	Person\end{tabular} & \begin{tabular}[c]{@{}c@{}}\scriptsize Third-Person\\\scriptsize Singular\end{tabular} & \begin{tabular}[c]{@{}c@{}}\scriptsize Third-Person\\\scriptsize Plural\end{tabular} & \begin{tabular}[c]{@{}c@{}}Impersonal\\Pronouns\end{tabular} \\
        \midrule
     & Human (\citeyear{kacewicz2014pronoun}) & $1.00$ & $\g0.50$ & $1.30$ & \hspace{-0.77em}$-0.50$ & \hspace{-0.77em}$\g-0.20$ & $\g0.02$ & $\g0.02$ & $\g0.40$ \\
      \multicolumn{10}{l}{\grayc \textit{Qwen 2.5 0.5B Instruct}}\\
      & Prompt 1 & $\g 0.49$ & \hspace{-0.77em}$\g-1.27$ & \hspace{-0.77em}$\g-0.21$ & \hspace{-0.77em}$\g-0.47$ & \hspace{-0.77em}$\g-0.07$ & \hspace{-0.77em}$\g-0.01$ & \hspace{-0.77em}$\g-0.51$ & $ 0.08$ \\
      & Prompt 2 & $\g 1.40$ & \hspace{-0.77em}$\g-0.22$ & \hspace{-0.77em}$\g-0.05$ & $\g 0.02$ & \hspace{-0.77em}$\g-0.16$ & $\g 0.00$ & \hspace{-0.77em}$\g-0.02$ & \hspace{-0.77em}$\g-0.19$ \\
      & Prompt 3 &$\g 0.28$ & \hspace{-0.77em}$\g-0.83$ & \hspace{-0.77em}$\g-0.04$ & \hspace{-0.77em}$\g-0.50$ & \hspace{-0.77em}$\g-0.13$ & \hspace{-0.77em}$\g-0.00$ & \hspace{-0.77em}$\g-0.16$ & \hspace{-0.77em}$\g-0.07$\\
      & Prompt 4 & $\g 0.08$ & $\g 0.08$ & \hspace{-0.77em}$\g-0.03$ & $\g 0.10$ & $\g 0.11$ & $\g 0.00$ & \hspace{-0.77em}$\g-0.11$ & \hspace{-0.77em}$\g-0.05$\\
    \multicolumn{10}{l}{\grayc \textit{Qwen 2.5 1.5B Instruct}}\\
    & Prompt 1 & \hspace{-0.77em}$\g-0.31$ & $\g 0.13$ & \hspace{-0.77em}$\g-0.05$ & $\g 0.32$ & \hspace{-0.77em}$\g-0.18$ & \hspace{-0.77em}$\g-0.01$ & $\g 0.05$ & $\g 0.06$\\
    & Prompt 2 & \hspace{-0.77em}$-0.70$ & $ 1.06$ & \hspace{-0.77em}$\g-0.02$ & $ 0.90$ & $ 0.01$ & $\g 0.00$ & $\g 0.17$ & $\g 0.06$ \\
    & Prompt 3 & \hspace{-0.77em}$\g-0.58$ & $\g 0.74$ & $\g-0.05$ & $\g 0.72$ & \hspace{-0.77em}$\g-0.00$ & \hspace{-0.77em}$\g-0.01$ & $\g 0.08$ & \hspace{-0.77em}$\g-0.07$ \\
    & Prompt 4 & \hspace{-0.77em}$\g-0.21$ & $\g 0.39$ & $\g 0.01$ & $\g 0.29$ & $\g 0.01$ & \hspace{-0.77em}$\g-0.01$ & $\g 0.09$ & \hspace{-0.77em}$\g-0.00$\\
    \multicolumn{10}{l}{\grayc \textit{Qwen 2.5 3B Instruct}}\\
    & Prompt 1 & $\g 0.93$ & $ 1.12$ & \hspace{-0.77em}$\g-0.37$ & $\g 0.38$ & $ 0.98$ & $ 0.07$ & $ 0.06$ & \hspace{-0.77em}$\g-0.16$ \\
    & Prompt 2 & $\g 0.71$ & $\g 0.15$ & \hspace{-0.77em}$-0.55$ & $\g 0.26$ & $ 0.35$ & \hspace{-0.77em}$\g-0.02$ & $ 0.11$ & \hspace{-0.77em}$-0.51$ \\
    & Prompt 3 & $\g 1.28$ & $ 0.81$ & \hspace{-0.77em}$\g-0.50$ & \hspace{-0.77em}$\g-0.19$ & $ 1.44$ & $\g 0.07$ & \hspace{-0.77em}$\g-0.01$ & \hspace{-0.77em}$\g-0.36$ \\
    & Prompt 4 & $\g 0.97$ & $ 1.34$ & \hspace{-0.77em}$\g-0.54$ & $ 0.92$ & $ 0.71$ & $\g 0.03$ & $\g 0.21$ & \hspace{-0.77em}$\g-0.34$ \\
    \multicolumn{10}{l}{\grayc \textit{Qwen 2.5 7B Instruct}}\\
    & Prompt 1 & \hspace{-0.77em}$-0.87$ & $ 1.34$ & $ 0.93$ & $ 0.81$ & \hspace{-0.77em}$\g-0.07$ & \hspace{-0.77em}$\g-0.00$ & \hspace{-0.77em}$\g-0.32$ & $ 0.28$ \\
    & Prompt 2 & $ 0.79$ & \hspace{-0.77em}$-0.80$ & \hspace{-0.77em}$-0.79$ & $\g 0.07$ & \hspace{-0.77em}$\g-0.08$ & $ 0.03$ & \hspace{-0.77em}$\g-0.03$ & $\g 0.01$ \\
    & Prompt 3 & \hspace{-0.77em}$\g-0.07$ & $\g 0.32$ & \hspace{-0.77em}$\g-0.11$ & $\g 0.12$ & $ 0.21$ & $ 0.05$ & $\g 0.06$ & \hspace{-0.77em}$\g-0.03$ \\
    & Prompt 4 & $ 1.36$ & \hspace{-0.77em}$-0.73$ & \hspace{-0.77em}$\g-0.33$ & \hspace{-0.77em}$-0.61$ & $ 0.20$ & $ 0.05$ & $\g-0.04$ & \hspace{-0.77em}$\g-0.13$\\
    \multicolumn{10}{l}{\grayc \textit{Qwen 2.5 14B Instruct}}\\
    & Prompt 1 & \hspace{-0.77em}$-0.47$ & $ 1.00$ & $\g 0.33$ & $ 0.53$ & $\g 0.19$ & \hspace{-0.77em}$\g-0.02$ & \hspace{-0.77em}$\g-0.02$ & $\g 0.08$\\
    & Prompt 2 & \hspace{-0.77em}$\g-0.72$ & $\g 0.67$ & \hspace{-0.77em}$-0.40$ & $ 0.66$ & $ 0.53$ & \hspace{-0.77em}$\g-0.01$ & \hspace{-0.77em}$\g-0.11$ & $\g 0.05$ \\
    & Prompt 3 & \hspace{-0.77em}$-0.60$ & $ 1.42$ & $\g 0.20$ & $ 0.64$ & $ 0.88$ & \hspace{-0.77em}$\g-0.03$ & \hspace{-0.77em}$\g-0.27$ & $\g 0.12$ \\
    & Prompt 4 & $\g 0.64$ & \hspace{-0.77em}$\g-0.28$ & \hspace{-0.77em}$-0.34$ & $\g 0.17$ & $ 0.14$ & $ 0.01$ & \hspace{-0.77em}$-0.26$ & \hspace{-0.77em}$\g-0.06$\\
    \multicolumn{10}{l}{\grayc \textit{Qwen 2.5 32B Instruct}}\\
    & Prompt 1 & \hspace{-0.77em}$-0.21$ & $ 0.68$ & \hspace{-0.77em}$\g-0.13$ & $ 0.50$ & $ 0.32$ & \hspace{-0.77em}$-0.01$ & \hspace{-0.77em}$\g-0.01$ & \hspace{-0.77em}$\g-0.13$ \\
    & Prompt 2 & \hspace{-0.77em}$-1.24$ & $ 1.43$ & \hspace{-0.77em}$\g-0.12$ & $ 1.06$ & $ 0.43$ & \hspace{-0.77em}$\g-0.04$ & \hspace{-0.77em}$\g 0.10$ & $\g 0.14$ \\
    & Prompt 3 & \hspace{-0.77em}$-0.85$ & $ 1.42$ & $\g 0.05$ & $ 1.01$ & $ 0.57$ & \hspace{-0.77em}$-0.02$ & \hspace{-0.77em}$\g-0.19$ & \hspace{-0.77em}$\g-0.05$ \\
    & Prompt 4 & \hspace{-0.77em}$-1.41$ & $ 1.85$ & \hspace{-0.77em}$\g-0.11$ & $ 1.38$ & $ 0.77$ & \hspace{-0.77em}$\g-0.03$ & \hspace{-0.77em}$\g-0.17$ & \hspace{-0.77em}$\g-0.07$\\
    \multicolumn{10}{l}{\grayc \textit{Qwen 2.5 72B Instruct}}\\
    & Prompt 1 & \hspace{-0.77em}$\g-0.09$ & $ 0.83$ & \hspace{-0.77em}$\g-0.16$ & $ 0.62$ & $ 0.56$ & \hspace{-0.77em}$\g-0.03$ & \hspace{-0.77em}$-0.15$ & \hspace{-0.77em}$-0.43$ \\
    & Prompt 2 & $ 0.21$ & $ 2.09$ & $\g 0.10$ & $ 0.66$ & $ 1.84$ & \hspace{-0.77em}$\g-0.02$ & \hspace{-0.77em}$-0.48$ & \hspace{-0.77em}$-0.66$ \\
    & Prompt 3 & $\g 0.01$ & $ 0.87$ & \hspace{-0.77em}$-0.39$ & $ 0.63$ & $ 0.86$ & $\g 0.00$ & \hspace{-0.77em}$-0.22$ & \hspace{-0.77em}$-0.40$ \\
    & Prompt 4 & \hspace{-0.77em}$-0.27$ & $ 1.06$ & \hspace{-0.77em}$\g-0.07$ & $ 0.72$ & $ 0.47$ & \hspace{-0.77em}$\g-0.02$ & \hspace{-0.77em}$\g-0.03$ & \hspace{-0.77em}$\g-0.20$ \\
    \bottomrule
    \end{tabular}
    }
    \caption{
    Results for LLM agents based on QWen models corresponding to \Cref{fig: simple-agent-qwen-family}.}
    \label{tab:qwen-family-studies}
\end{table*}

\Cref{tab:aggregated} presents the complete pronoun usage difference results corresponding to \Cref{fig: simple-agent-advanced-models} and \Cref{fig: complex-agents}. 
We can see that neither the LLM agents from different model families nor the specialized agents can consistently elicit human-like pronoun usage in their interaction process.
In most cases, their patterns contradict human patterns.

\Cref{tab:gpt-family-studies,tab:llama-family-studies,tab:qwen-family-studies} present the results for models from GPT family (corresponding to \Cref{fig: simple-agent-gpt-family}), Llama family (corresponding to \Cref{fig: simple-agent-llama-family}) and QWen family (corresponding to \Cref{fig: simple-agent-qwen-family}), respectively.
We observe that there are model-specific patterns across different prompts.
For instance, leader agents tend to use first-person singular pronouns more often across Llama 3.1 8B models, while the first-person singular pronoun usage difference is not statistically significant across all the Qwen 2.5 models.

\subsection{Other Effects}
\label{app-subsec: other-effects}

\begin{figure}[t]
    \centering
    \begin{subfigure}{\linewidth}
        \includegraphics[width=0.95\linewidth]{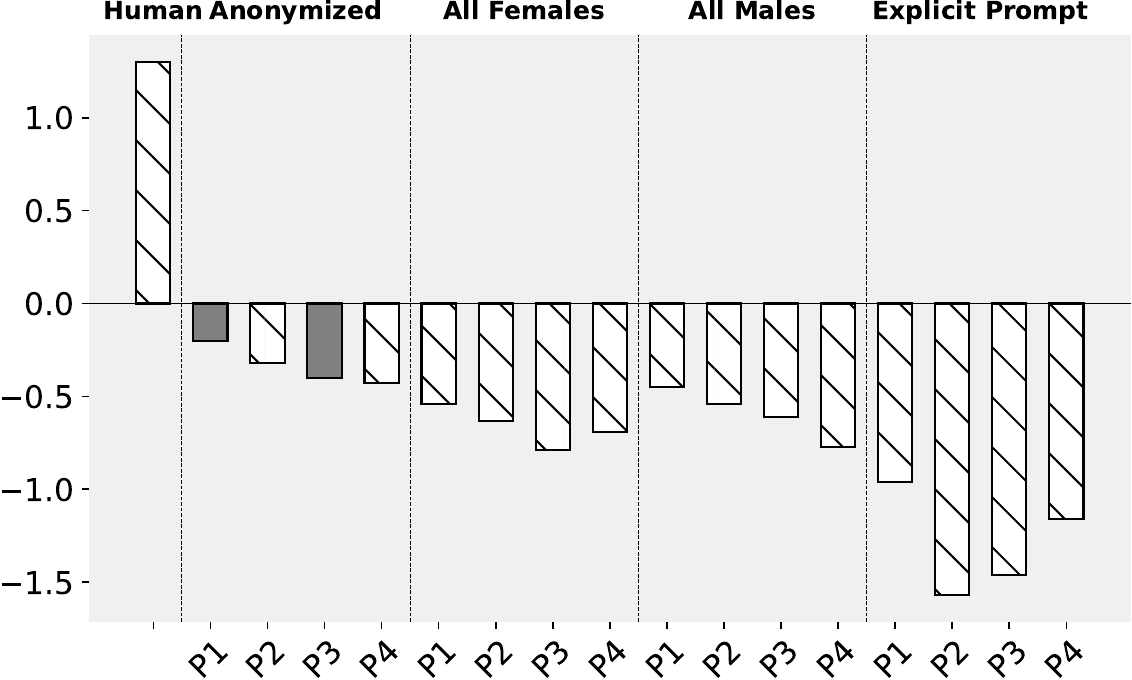}
        \caption{First-person singular pronouns.}
        \label{fig:first-person-singular-others}
    \end{subfigure}
    \begin{subfigure}{\linewidth}
        \includegraphics[width=0.95\linewidth]{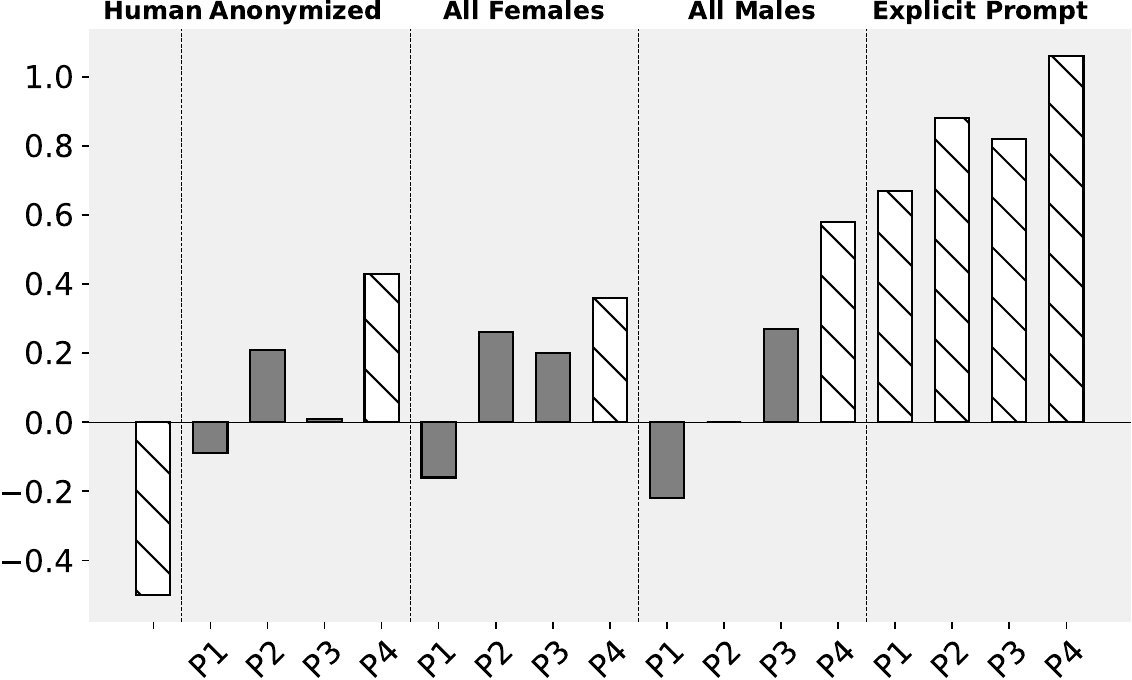}
        \caption{First-person plural pronouns.}
        \label{fig:first-person-plural-others}
    \end{subfigure}   
    \caption{Differences in pronoun usage when names are anonymized, all agents are assigned female or male genders or an explicit prompt is used (described in \Cref{app-subsec: other-effects}).
    We conduct the experiments based on GPT-4o.
    None of these adjustments can elicit human-like pronoun usage patterns.}
    \label{fig:pronoun-other-effects}
\end{figure}

We explore the effects of various settings based on the four prompts in \Cref{tab:persona-prompts} using GPT-4o and report our results for the first-person singular and plural pronouns in \Cref{fig:pronoun-other-effects}.
For the setting of ``Anonymized'', we anonymize the names of agents by names such as ``Name 1'', ``Name 2'', etc.
For the settings of ``All Females'' or ``All Males'', we assign the genders of all the LLM agents as females or males, respectively.
For the setting of ``Explicit Prompt'', we add a sentence of ``Please use first-person plural forms (e.g. we, us) more often'' if the agent acts as leaders, and ``Please use first-person singular forms (e.g. I, me) more often'' if the agent acts as non-leaders.
We highlight that none of these adjustments can elicit human-like pronoun usage patterns.

\begin{table}[t]
    \small
    \centering
    \begin{tabular}{cp{17em}}
        \toprule
        \textit{Role} & \multicolumn{1}{c}{\textit{Utterance}} \\
        \midrule
        \multirow{2}{*}{Leader} &  I will start by sharing my initial thoughts, and then I’d love to hear from each of you.\\
        \midrule
        \multirow{5}{*}{Non-leader} & Hi everyone, thanks for bringing this up. 
I think it’s really important that we all share our perspectives on how these strategies can positively contribute to both our service quality and company culture at Jupiter. \\
        \bottomrule
    \end{tabular}
    \caption{Example of the utterance from leader agent versus non-leader agent.
    This example corresponds to GPT-4o with Prompt 1.}
    \label{tab:agent-comparison}
\end{table}

\subsection{Examples of Model Responses}
\label{app-subsec: model-response-examples}

We observe that there is minimal distinction in language use when the model assumes the role of a leader versus a non-leader.
For instance, when the GPT-4o assumes the role of leader and non-leader in \Cref{tab:agent-comparison}, both responses exhibit a collaborative and inclusive tone, but without a clear distinction between the roles.



\section{Dialogue Examples}
\label{app-sec:dialogue-example}

\Cref{tab:transcript-comparisons} presents an example from the original human study by \citet{kacewicz2014pronoun} and an example from the dialogue by LLM agents using GPT-4o with Prompt 1.

\begin{table*}[t]
    \small
    \centering
    \begin{tabular}{cp{40em}}
    \toprule
    \multicolumn{2}{l}{\grayc Human Transcripts \citep{kacewicz2014pronoun}} \\
    Person 1     &  Okay. Let's work on the rankings.  How do you guys want to go through and do this?  Do you want to...shall we just go through and talk about how important each one is and then rank them?  Because it's kind of hard to just decide which one is number 1.\\
    Person 2     & Yeah. \\
    Person 3     & Yeah. \\
    Person 4    & Yeah I think that's a good idea. \\
    Person 1    & Alright.  Alright, we'll rank things 1 to 5 on how important they are.  I guess 1 will be most important and 5 the least.  And then we'll just have the groups... \\
    Person 2    & Sounds good. \\
    ...\\
    \bottomrule
    \multicolumn{2}{l}{\grayc GPT-4o (Prompt 1) Transcripts} \\
    Agent 1 & Hi everyone! I hope you're all doing well. Today we need to discuss and rank these strategies for improving service quality based on their contribution to our company culture. I've already thought about this a bit and have a proposed order, but I'm eager to hear your thoughts as well. Here’s my suggested ranking... \\
    Agent 2 & Thanks for your input. You’ve made some great points, and I appreciate the thought you’ve put into your ranking. I agree that training our customer service representatives should be at the top of our list. It's the foundation for everything else we want to achieve. 
I’d like to hear from the rest of the team before we finalize our ranking. What are your thoughts on this? How would you prioritize these items? \\
    Agent 3 & Thanks for sharing your thoughts. I think you’ve highlighted some very important points. I agree that training our customer service representatives should be a top priority because it directly impacts their ability to handle customer interactions effectively... \\
    ... \\
    \bottomrule
    \end{tabular}
    \caption{Transcripts from human studies by \citet{kacewicz2014pronoun} and transcripts from GPT-4o with Prompt 1.}
    \label{tab:transcript-comparisons}
\end{table*}


\end{document}